%% file: paper.tex
\documentclass[11pt]{article}

\usepackage[final]{acl}

\usepackage{times}
\usepackage{latexsym}

\usepackage[T1]{fontenc}

\usepackage[utf8]{inputenc}

\usepackage{microtype}

\usepackage{inconsolata}

\usepackage{graphicx}
\usepackage{amsmath}
\usepackage{booktabs}
\usepackage{adjustbox}

\title{Memory {vs.} Context? Influential Factors of Factual Recall in Language Models}

\author{Guilhem Fouilhé \\
  IRIT  \\
  ANITI  \\
  \texttt{g.g.a.fouilhe@gmail.com} \\\And
  Nicholas Asher \\
  IRIT  \\
  CNRS \\
  \texttt{nicholas.asher@irit.fr} \\\And
  Philippe Muller \\
  IRIT \\
  Université de Toulouse  \\
  \texttt{philippe.muller@irit.fr} \\}

\begin{document}
\maketitle
\begin{abstract}
  We reproduce and stress-test the work of~\citet{yu-etal-2023-characterizing}, who characterize how language models (LMs) arbitrate between memorized knowledge and contradictory in-context statements.   We replicate their world-capitals experiments on 31 models spanning Pythia, GPT-2, Qwen3, and Ministral families, including base and post-trained variants, and extend evaluations to five additional knowledge relation types from the ParaConflict dataset.  We empirically confirm most of their original findings: larger models and higher-frequency entities tend to favor memorized answers, with substantial family-level variance.
  However, several conclusions do not generalize cleanly: entity-frequency effects disappear on Qwen3-14B and 32B; post-training shifts the memory-context trade-off inconsistently across families; question phrasing alone can change a model's reliance on memorized knowledge by up to 80 percentage points; and semantically unrelated prose can mimic coherent supporting context.
  Our results clarify where Yu et al.'s claims hold and to what extent they generalize to other prompts.\footnote{Our code is available at \\ \url{https://github.com/gfouilhe/IC-factual-public}}
\end{abstract}

\section{Introduction}

Language Models (LMs) based on the Transformer architecture \cite{vaswani2017attention} show evidence of factual recall through memorization of the training data \cite{meng2022locating}, meaning they can be prompted to output factual knowledge. For instance, a generative model prompted with ``{\it The capital of Hungary is}'' is likely to predict {\it Budapest} as the next token(s).  This capacity scales with model size \cite{ICLR2025_26d3c9a6} and topic frequency \cite{smith2026predictable} (the frequency of a fact in the training data).%

This capacity, however, sometimes conflicts with another core capacity of LMs, \textit{In-Context Learning} or ICL \cite{brown2020language}, which enables models to use information from a prompt's context to make predictions.  This feature is particularly useful because
facts can change.
For instance, the statement \textit{The Argentina men's national football team are the world champions} was true between 2022 and 2026, but became obsolete after the 2026 final, causing models' memorized knowledge to be outdated. Model prompting can be used to explicitly introduce new facts (e.g.\ adding \textit{Spain defeated Argentina in the 2026 World Cup final} at the beginning of the prompt).
In such cases, it is usually desirable that models give priority to the context over memorized knowledge. But it is not always the case: we also want models to be robust against false information in prompt injections. Understanding how models behave in these situations is important for understanding their limitations and improving their robustness.

\citet{yu-etal-2023-characterizing} show that different models (ranging from Pythia-70M to Pythia-2.8B and GPT-2 to GPT-2-XL) have different ways of handling prompts of the form \textit{``The capital of Poland is London. Q: What is the capital of Poland? A:''}, designed to minimally introduce a conflict between memorized knowledge and context. The mechanism each model uses depends both on the frequency of the subject of the fact (the country, e.g., Poland) and the frequency of the overwriting information (the counterfactual city, e.g., London): as the frequency goes up, models are more likely to predict the memorized answer.

However, \citet{yu-etal-2023-characterizing} do not clarify how these results depend on the model architecture, size, and stage of training. Moreover, they establish the observed effects only on a single template and a single specific dataset (world-capitals).

\begin{figure*}[t]
  \centering
  \includegraphics[width=0.9\textwidth]{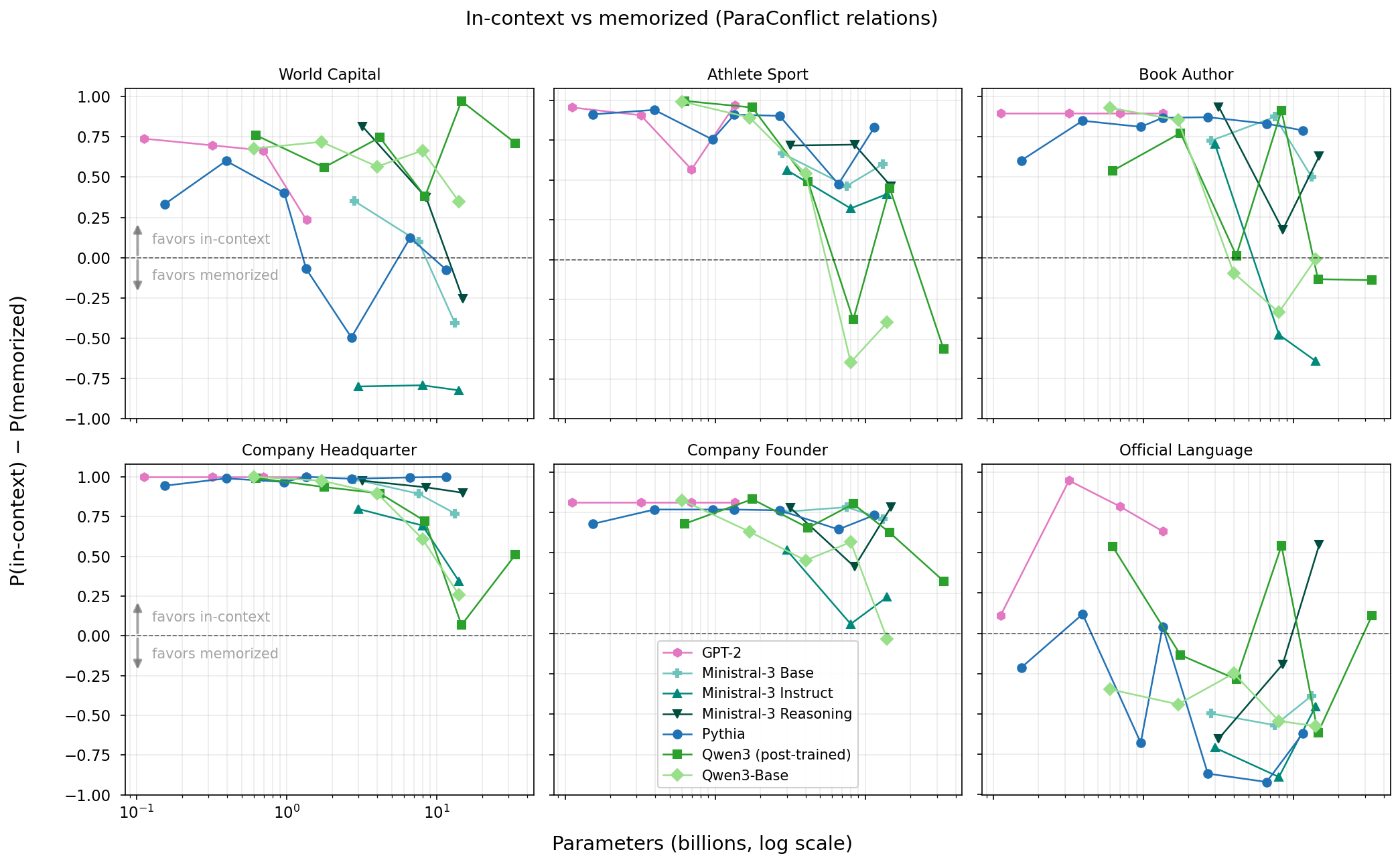}
  \caption{Comparison of several models' behavior with context/memorized knowledge conflict
    across parameter counts. Scores are differences in probability between in-context information and memorized information, averaged over instances of 6 ParaConflict relations. Positive scores means the model chooses in-context responses more often.}
  \label{fig:context_memory_score_gen_all}
\end{figure*}

These limitations provide the motivation for our paper: we re-evaluate those experiments on more models, varying prompt-design factors and testing the robustness of the results on more varied datasets.  We establish \citet{yu-etal-2023-characterizing}'s result that larger models and higher-frequency entities tend to favor memorized answers, with substantial family-level variance on a broader range of topics and across 31 models.  We also show that: topic frequency effects vary across some models; post-training shifts the memory-context trade-off inconsistently across model families; question phrasing in the prompt can substantially alter a model's reliance on memorized knowledge, as can the use of semantically unrelated prose.

\section{Related Work}
\input{related_work.tex}
\section{Methods}
\input{methods.tex}

\begin{figure*}[t]
  \centering
  \includegraphics[width=\textwidth]{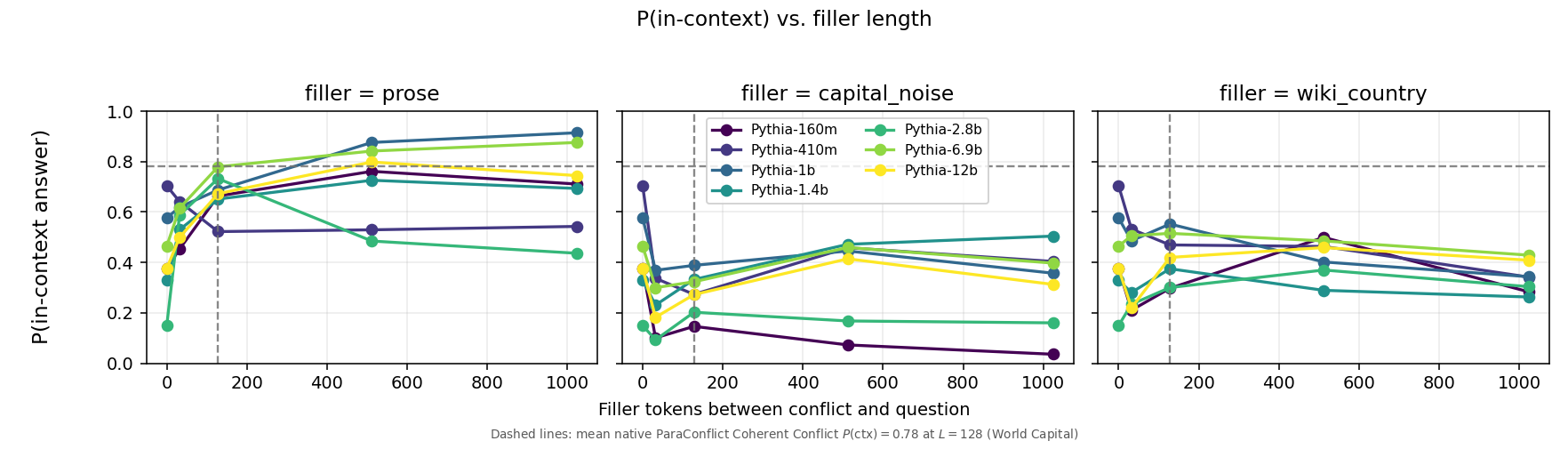}
  \caption[Effect of filler length and filler type across Pythia sizes]{$P(\text{in-context})$ vs.\ filler length for different fillers across Pythia sizes.}
  \label{fig:exp22-gen-length}
\end{figure*}

\section{Results}

\input{results.tex}

\section{Discussion}

Our experiments support five main findings:
\begin{enumerate}
  \item Model size is positively correlated with reliance on memorized knowledge, but the effect is non-linear and strongly family-dependent (e.g.\ it is stronger for Qwen3 than for the other families studied).
  \item Subject frequency in the training data is positively correlated with $P(\text{mem})$ for most models, reinforcing \citet{yu-etal-2023-characterizing}; however, the correlation vanishes for some models, notably Qwen3-14B and 32B.
  \item Post-training shifts the memory--context trade-off substantially but inconsistently: Qwen3 post-training and Ministral-3 Reasoning generally decrease $P(\text{mem})$, while Ministral-3 Instruct increases it, with per-size reversals of up to 60\,pp on individual relations.
  \item The question template alone can change $P(\text{mem})$ by up to 95\,pp at fixed model and conflict statement across all modern families tested (Figure~\ref{fig:cross-family-phrasing-swings}).
  \item A coherent passage supporting the context raises $P(\text{in-context})$ on every model tested; on Pythia, semantically unrelated prose of the same length has a comparable effect, but across other families unrelated prose exhibits inconsistent effects, sometimes shifting strongly in the opposite direction (Figure~\ref{fig:cross-family-prose-l128-delta}).
\end{enumerate}

Taken together, findings 3--5 suggest that memory--context arbitration is not a stable, model-level property: within a single model, question phrasing, filler content, and conflict position can swing the outcome by tens of percentage points. Characterizations based on a single template and dataset therefore risk overstating generality.

Besides model specificity, Qwen3 counterexamples to the frequency effect admit two readings. Models trained on far more data may store even rare entities robustly enough that frequency ceases to discriminate, making the effect a signature of undertrained knowledge. Alternatively, our \texttt{wordfreq} proxy may diverge from the actual pretraining distribution---a limitation shared by any frequency analysis of closed-data models.

\section{Conclusion}

We reproduced and extended \citet{yu-etal-2023-characterizing}'s study of factual recall under substitution conflicts across 31 models, six ParaConflict relation types, and targeted prompt ablations. Most original trends hold at scale: larger models and higher-frequency entities tend to favor memorized answers. Nevertheless several conclusions are prompt- and family-sensitive: entity-frequency effects vanish on some Qwen3 sizes, post-training shifts the memory--context trade-off inconsistently, question phrasing alone can swing $P(\text{mem})$ by up to 95 percentage points, and semantically unrelated filler can mimic coherent supporting context. A cross-family logit-lens analysis reveals that preference for memorized versus in-context answers becomes directly decodable primarily in late network layers. Together, these findings clarify the boundaries of prior claims and establish prompt design as a decisive factor in memory--context arbitration. Extending our ablation framework to broader conflict benchmarks such as ConflictBank \citep{su2024conflictbank} is a natural next step.

\section*{Limitations}
Despite extending the original study, there are still restrictions to the present work, along multiple dimensions:
(a) conflicts investigated here are limited to conflicts between in-context information and memorized knowledge by the model; internal model or context conflicts might be elicited in comparable ways or demand different methodologies.
(b) the type of conflict is factual, assuming a universal stable ground truth, and ignore e.g. temporal aspects (world capitals can change, but are relatively stable in general, at least in the available written data that feed current model training.)
(c) we show a lot of variance associated to certain factors, for instance phrasing of the conflict, but it remains unclear what general linguistic concepts are responsible for the variations.
Finally, the approach is limited to models for which we can access prediction probabilities, so currently only open-weight
models, and for practical reasons we capped the model sizes.

\section*{Acknowledgments}

This work was supported by the Artificial and
Natural Intelligence Toulouse Institute (ANITI). ANITI is funded by the France 2030 program under the
Grant agreement ANR-23-IACL-0002.
This research is also supported by the National Research Foundation, Prime Minister’s Office, Singapore under its Campus for Research Excellence and Technological Enterprise (CREATE) programme.
Computational resources were provided by a French government grant managed by the Agence Nationale de la Recherche under the "Investissements d'avenir" program (reference "ANR-21-ESRE-0051").

\bibliography{custom}

\appendix
\input{appendix.tex}

\end{document}

%% file: related_work.tex
\citet{yu-etal-2023-characterizing} characterize how LMs arbitrate between memorized knowledge and in-context contradictions; \citet{li2025taming} extend this line of work with the ParaConflict benchmark and a mechanistic analysis of attention heads.
They find that ``context heads'' and ``memory heads'' can switch roles when the degree of conflict changes slightly, so \citet{yu-etal-2023-characterizing}'s observation that some heads promote context or memorized knowledge does not generalize well and intervening on a single head may not robustly steer models toward either source.
They also show that multiple interventions are not necessarily additive, and propose a method %
to improve intervention robustness.

\citet{longpre-etal-2021-entity} formalize the contextual--parametric knowledge conflict and provide a framework for evaluating model behavior; the setup of \citet{yu-etal-2023-characterizing} is essentially what they call \emph{corpus substitution}.
They also propose a mitigation technique based on continued training.

\citet{xie2024adaptive} introduce a framework to systematically construct counterevidence to the parametric memory of LMs.
They find that model behavior depends strongly on coherently presented in-prompt evidence.

The context--memory conflict we study is one kind of \emph{knowledge conflict}. \citet{xu-etal-2024-knowledge-conflicts} taxonomize knowledge conflicts into three categories: context--memory conflict (context contradicting parametric knowledge), \emph{inter-context} conflict (contradictions among different pieces of context, e.g.\ noisy or misinformed retrieved documents in RAG settings), and \emph{intra-memory} conflict (inconsistencies within the model's own parametric knowledge, surfacing as divergent answers to differently phrased queries).
While we focus on the first category, the analysis we provide could also benefit the study of other types of conflicts.

%% file: methods.tex
\subsection{Task formulation}

We investigate LM behavior when prompted with a \emph{substitution conflict}: a counterfactual
statement followed by a query for the same fact. We use the template
from \citet{yu-etal-2023-characterizing} unless noted otherwise:
\begin{quote}
  \small
  \texttt{The capital of \{country\} is \{distractor\}. Q: What is the capital of \{country\}? A:}
\end{quote}
For each prompt the model must choose between the memorized capital
$c_{mem}$ and the in-context distractor $c_{ctx}$. We report
\\ $P(\text{in-context}) = P(c_{ctx} \mid prompt)$ and \\ $P(\text{memorized}) = P(c_{mem} \mid prompt)$.

\subsection{Experimental design}

\paragraph{Datasets}
We build prompts from the ParaConflict Dataset \cite{li2025taming}.
The \textsc{World Capital} category is essentially equivalent to the dataset used by \citet{yu-etal-2023-characterizing}, but only provides one distractor per country. We extend it following \citet{yu-etal-2023-characterizing}: for every country with gold capital $c_{mem}$ and every other country's capital $c'_{mem}$, we set $\{distractor\} := c'_{mem}$ in the template. We omit pairs where $c'_{mem}$ is also a correct answer for the queried country. This yields $n{=}47{,}299$ prompts covering 218 countries.

We also study five additional ParaConflict categories (\textsc{Athlete Sport} (n=532), \textsc{Book Author} (n=500), \textsc{Company Headquarter} (n=500), \textsc{Company Founder} (n=203), \textsc{Official Language} (n=193)) to test relation specificity.
However, the main focus of this paper is on the \textsc{World Capital} category, because it is the most studied in prior work.

\paragraph{Context-length and filler tokens.}

\citet{yu-etal-2023-characterizing} focused on substitution conflicts of the form \texttt{conflict|\,question} which fails to capture the wide range of contexts in which these conflicts can arise.
To investigate the effect of adding coherent context before the question, ParaConflict provides a \texttt{Coherent Conflict} variant: a fluent passage of $\approx$120 tokens that elaborates the counterfactual claim. Across all 31 models we evaluate, replacing the substitution template with this coherent passage raises $P(\text{in-context})$ on every model (median $\Delta = +0.19$; range $+0.01$--$+0.57$; Appendix~\ref{sec:appendix-native-sub-vs-coh}, Table~\ref{tab:paraconflict-native-sub-vs-coh}). In the same direction, coherent supporting context can strongly steer models toward the in-context answer \citep{xie2024adaptive}.

To isolate this effect, we insert a configurable
filler \emph{between} the conflict statement and the question while
holding the conflict and query fixed. Filler length is measured in
\emph{tokens} under the model's tokenizer and tiled or truncated to
the target length. Unless stated otherwise, the conflict sits in
\emph{tail} position (\texttt{[conflict] [filler] [question]}, i.e., separated from the question by filler); we
also test \emph{front} (\texttt{[filler] [conflict] [question]}, immediately before the question) and \emph{middle} (\texttt{[½filler] [conflict] [½filler] [question]}) placements at selected
lengths. We use the following filler types:
\begin{itemize}
  \item \textbf{prose} --- a fixed, AI-generated English
        paragraph (plant-cell biology) with no country or capital names;
  \item \textbf{capital\_noise} --- per-prompt sequences of true
        statements \texttt{The capital of X is Y.} sampled from the
        country pool, excluding the queried country and the distractor's country;
  \item \textbf{capital\_noise\_false} --- same surface form as
        \textbf{capital\_noise} but with shuffled (false) country--capital
        pairings. Queried country's capital and distractor capital are also excluded;
  \item \textbf{wiki\_*} --- Wikipedia country descriptions
        (for the queried \texttt{country}, the \texttt{distractor\_country} whose actual capital is the distractor, or an \texttt{unrelated\_country}) with the relevant capital
        redacted;
\end{itemize}
We truncate context lengths to $L \in \{0, 32, 128, 512, 1024\}$ tokens as closely as possible while maintaining correct sentence structure.

\paragraph{Template and conflict ablations.}
\label{sec:methods-template-ablations}
We vary the question template: \texttt{qa} template from \cite{yu-etal-2023-characterizing} vs.\
\texttt{bare} continuation (\texttt{"The capital of \{country\} is"}),
plus \texttt{possessive} (\texttt{"\{country\}'s capital is"}), \texttt{learned} (\texttt{"I learned that the capital of \{country\} is"}), and \texttt{of\_course} (\texttt{"Of course, the capital of \{country\} is"})
variants;  the conflict template: \texttt{standard} vs.\
\texttt{negated}: \texttt{"...is not \{distractor\}."}; and the number of
conflict copies $K \in \{1,2,4,8\}$ at $L{=}0$.
Full results for the alternative question templates are in
Appendix~\ref{sec:appendix-alt-scaffolds}; negated-conflict and
conflict-copy ablations are in
Appendices~\ref{sec:appendix-negated-conflict}
and~\ref{sec:appendix-repetition-sweep}.

\subsection{Models}

\paragraph{Pythia.}
We evaluate the full EleutherAI Pythia suite (160M--12B) \cite{pythia}, including
6.9B and 12B models beyond the sizes studied in
\citet{yu-etal-2023-characterizing}. All Pythia models are trained on 300B tokens.

\paragraph{GPT-2.}
We include the OpenAI GPT-2 series \cite{gpt2}: \texttt{small} (124M),
\texttt{medium} (355M), \texttt{large} (774M), and \texttt{xl}
(1.5B). The training data size is not disclosed.

\paragraph{Qwen3.}
We also include Qwen3 \cite{yang2025qwen3} dense models at 0.6B,
1.7B, 4B, 8B, 14B, and 32B (post-trained, instruction-tuned
models) and matching Qwen3-Base pretrained models from
0.6B through 14B (no public 32B-Base release). Comparing base vs.\ post-trained isolates the effect of
post-training from pretraining scale and architecture. All Qwen3 models are reportedly trained on 36T tokens.

\paragraph{Ministral-3.}
We evaluate Mistral's Ministral-3 \cite{liu2026ministral} dense models at 3B, 8B, and
14B in three post-training stages: \texttt{Base}, \texttt{Instruct},
and \texttt{Reasoning}. Ministral-3 models are reportedly trained on 1T to 3T tokens.

\subsection{Answer classification}

For each prompt we label a response as \textbf{memorized}  if the model
favors the gold capital and  \textbf{in-context} if it
favors the distractor.

\paragraph{Generative classifier.}
Following \citet{yu-etal-2023-characterizing}, we decode
up to 10 new tokens greedily and classify by substring match against
gold aliases and the distractor.

\paragraph{Log-probability classifier.}
We verified the correlation of the generative classifier with a more direct evaluation that scores the two candidates by
log-probability of the continuation \texttt{ \{city\}} (leading space
included) given the prompt:
\[
  \hat{y} = \arg\max_{y \in \{c_{mem},\, c_{ctx}\}} \log P(y \mid \text{prompt}),
\]
where $c_{mem}$ is the memorized capital and $c_{ctx}$ the distractor. Unlike the 10-token generative classifier, this teacher-forced sequence score evaluates the complete target continuation and is immune to answer truncation and conversational scratchpad preambles (Appendix~\ref{sec:appendix-inference-specs}). The correlation between the generative and log-probability classifiers is high and discussed in Appendix~\ref{sec:appendix-generative-logprob-correlation}.

\subsection{Training-frequency proxy}
\label{sec:methods-training-freq}

\citet{yu-etal-2023-characterizing} relate memorization rates to
entities' frequency in the Pile \citep{gao2020pile}, Pythia's 825\,GB pretraining corpus. We approximate entity frequency with
the \texttt{wordfreq}\footnote{https://github.com/rspeer/wordfreq} package's Zipf score on a mixed web/Wikipedia/news
corpus. This proxy has the advantage of simplicity of reproduction and allows a fair comparison between models with open-access and private training data.
Scores are computed per country (subject) and per distractor country; leading articles
(\texttt{the}) are stripped so that \texttt{the United Kingdom} and
\texttt{France} are comparable. Full per-entity Zipf scores for all six
ParaConflict categories appear in Appendix~\ref{sec:appendix-paraconflict-freq}.
We bin subjects by Zipf and plot
$P(\text{memorized})$ vs.\ $P(\text{in-context})$ by bin, replicating
the paper's Figure 3 and parts of Figure 2.

\subsection{Mechanistic analysis: logit lens}
\label{sec:methods-logit-lens}

To inspect when the preference between memorized and in-context
answers becomes directly decodable in the residual stream, we apply the \emph{logit lens} \cite{logitlens}: for each prompt we run a forward
pass, project every
transformer block output through the model's final normalization and
vocabulary unembedding, and read off the log-probability of the
\emph{first token} of the memorized capital and of the distractor at
the last sequence position. We aggregate mean per-layer gaps
$\Delta_\ell = \log P(c_{ctx}) - \log P(c_{mem})$ over $n{=}500$ prompts per
condition on all models for the World Capital task.

%% file: results.tex
\subsection{Model sizes and memorization}
\label{sec:results-model-size}
While \citet{yu-etal-2023-characterizing} do not state this explicitly, their results suggest that bigger models tend to use memorized knowledge more than smaller models. %
We studied this phenomenon on all ParaConflict relation categories (Figure~\ref{fig:context_memory_score_gen_all}) and measured the correlation between model size and $P(\text{mem})$ (Table~\ref{tab:relations-size-mem}; by-family breakdown in Appendix), and the effect of post-training on $P(\text{mem})$ (Table~\ref{tab:relations-posttraining}; per-model breakdown in Appendix). Overall, we find that model size is significantly positively correlated with $P(\text{mem})$, despite high variance across model families and post-training stages. Post-training can either increase or decrease $P(\text{mem})$ depending on the model family and the relation category, with Qwen3 post-training generally decreasing $P(\text{mem})$ and Ministral-3 Instruct post-training generally increasing it.  Thus, post-training seems to have inconsistent effects across model families.

To determine whether low $P(\text{mem})$ under conflict reflects in-context reliance or simply an absence of memorized knowledge, we evaluate a baseline of factual recall without conflict using the bare continuation prompt ($P(\text{correct} \mid \text{clean prompt})$; e.g., \emph{"The capital of X is"}, \emph{"X plays the sport of"}; Figures~\ref{fig:clean-factual-recall} and \ref{fig:clean-factual-recall-lp} in Appendix~\ref{sec:appendix-clean-factual-recall}).
Across all 31 models and all six relations, baseline factual recall scales monotonically with parameter size (e.g., Pythia overall accuracy scales from 13\,\% at 160M to 53\,\% at 12B; Qwen3-Base from 28\,\% at 0.6B to 64\,\% at 14B; Ministral-3 Base reaches 78\,\%).
Importantly, this clean baseline reveals that the near-zero memorization rates observed under substitution conflict for certain relations (e.g., Book Author and Company Founder) in smaller models stem from an absence of parametric storage (0\,\% recall on Book Author for Pythia $\le$ 2.8B and GPT-2) rather than purely robust context following.
Conversely, Official Language and World Capital have high baseline recall ($\ge$85\,\% for models $\ge$1.4B), confirming that their behavior under conflict represents a genuine competition between memorized knowledge and in-context distractors.

\begin{table}[!ht]
    \centering
    \small
    \setlength{\tabcolsep}{4pt}
    \caption{Correlation of $P(\text{mem})$ with model size, pooled across all 31 models. We report the mean memorization rate $\bar{P}(\text{mem})$, Spearman $\rho$ between parameter count and $P(\text{mem})$, and two-sided $p$-values. Positive $\rho$ means larger models memorize more than smaller ones. $n$ is number of relation instances of each category.}
    \label{tab:relations-size-mem}
    \begin{tabular}{@{}lrrrr@{}}
        \toprule
        Category            & $n$    & $\bar{P}(\text{mem})$ & $\rho$ & $p$    \\
        \midrule
        World Capital       & 47,299 & 0.30                  & +0.48  & 0.006  \\
        Athlete Sport       & 532    & 0.22                  & +0.80  & <0.001 \\
        Book Author         & 500    & 0.18                  & +0.81  & <0.001 \\
        Company Headquarter & 500    & 0.07                  & +0.77  & <0.001 \\
        Company Founder     & 203    & 0.07                  & +0.81  & <0.001 \\
        Official Language   & 193    & 0.58                  & +0.43  & 0.016  \\
        \midrule
        All (pooled)        &        & 0.24                  & +0.53  & <0.001 \\
        \bottomrule
    \end{tabular}
\end{table}

\begin{table}[!ht]
    \centering
    \footnotesize
    \setlength{\tabcolsep}{4pt}
    \caption{Effect of post-training on $P(\text{mem})$. Positive $\Delta P(\text{mem})$ means post-training increases
        memorization.}
    \label{tab:relations-posttraining}
    \begin{tabular}{@{}lrrr@{}}
        \toprule
                            & \multicolumn{3}{c}{$\Delta P(\text{mem})$ vs.\ Base (pp)}                     \\
        \cmidrule(lr){2-4}
        Category
                            & Qwen3
                            & \shortstack{Ministral-3\\Instruct}
                            & \shortstack{Ministral-3\\Reasoning}                                           \\
        \midrule
        World Capital       & $-$10.0                                                   & $+$41.4 & $-$13.8 \\
        Athlete Sport       & $-$11.8                                                   & $+$7.1  & $-$3.0  \\
        Book Author         & $-$11.4                                                   & $+$38.0 & $+$4.5  \\
        Company Headquarter & $+$0.1                                                    & $+$12.1 & $-$2.4  \\
        Company Founder     & $-$11.1                                                   & $+$17.1 & $+$2.8  \\
        Official Language   & $-$31.8                                                   & $+$10.0 & $-$18.1 \\
        \midrule
        All (pooled)        & $-$12.7                                                   & $+$20.9 & $-$5.0  \\
        \bottomrule
    \end{tabular}
\end{table}

\subsection{Country frequency and memorization}

\label{sec:results-country-freq}

\citet{yu-etal-2023-characterizing} argue that, when context and memory
conflict, models should rely more on memorized answers for
\emph{high-frequency} queried countries, reflecting stronger factual
storage for entities seen more often during pretraining.  We test this
hypothesis on the full world-capitals dataset
across 31 models.  For each model we correlate per-country
\texttt{wordfreq} Zipf with $P(\text{mem})$ and confirm the trend with ten equal-count Zipf deciles.  The hypothesis is supported at least partially: in 17 of 31 models,
country-level Spearman $\rho \geq 0.3$ and the highest-frequency decile
exceeds the lowest by at least 5 percentage points; Pythia (except 160M),
GPT-2 medium--XL, and most Ministral variants follow the Yu et al.\
pattern.  Counterexamples notably include the largest model tested, Qwen3-32B ($\rho \approx 0$).
Further analyses are reported in Appendix~\ref{sec:appendix-country-freq} (Table~\ref{tab:country-freq-mem} and Figure~\ref{fig:country-freq-mem}).

\subsection{Context length and filler tokens}
\label{sec:results-context-sweep}

Figure~\ref{fig:exp22-gen-length} shows generative-classifier outcomes for
three representative fillers (\texttt{prose}, \texttt{capital\_noise},
\texttt{wiki\_country}), context lengths $L \in \{0, 32, 128, 512, 1024\}$,
and Pythia models (160M--12B; $n{=}2000$ prompts per cell).
For medium and large models (Pythia $\ge$1.4B), inserting unrelated \texttt{prose} filler between conflict and question sharply increases $P(\text{in-context})$ (e.g., from 29\% at $L{=}0$ to 93\% at $L{=}128$ for 2.8B), showing that even semantically inert text can mimic the effect of a coherent passage. In contrast, for smaller models (160M and 410M, the grey and green lines in the left panel), increasing filler length steadily dilutes context retention, pulling $P(\text{in-context})$ downward. Meanwhile, \texttt{capital\_noise} and \texttt{wiki\_country} suppress in-context answers across virtually all sizes as length grows.
The full six-filler grid is in Appendix~\ref{sec:appendix-exp22-full}.
Evaluating $L{=}128$ unrelated prose across all 31 models (Appendix~\ref{sec:appendix-cross-family-prose}, Figure~\ref{fig:cross-family-prose-l128-delta}) reveals that models outside Pythia exhibit inconsistent effects. While positive shifts appear on a few models (e.g., Qwen3-Base-4B: $+16.4$\,pp; Ministral-3 Reasoning-8B: $+22.8$\,pp), other families often move strongly in the opposite direction, with unrelated filler decreasing $P(\text{in-context})$ by up to $-28.2$\,pp in Qwen3 and $-38.8$\,pp in Ministral-3 as memorized priors re-assert themselves. Thus, unlike coherent supporting context which elevates $P(\text{in-context})$ universally (Table~\ref{tab:paraconflict-native-sub-vs-coh}), the boost from unrelated prose is not consistent across model families.
Conflict-position ablations (\texttt{front}/\texttt{middle}/\texttt{tail})
are in Appendix~\ref{sec:appendix-position}.
Question template ablations across all 31 models (Appendix~\ref{sec:appendix-alt-scaffolds}, Figure~\ref{fig:cross-family-phrasing-swings}) further reveal that prompt framing alone can swing $P(\text{mem})$ by up to 95.0\,pp (e.g., \texttt{qa} vs.\ \texttt{of\_course} in Qwen3-14B), demonstrating that phrasing sensitivity is architecture-general across modern models.

\subsection{Late-layer direct decodability: cross-family logit lens}
\label{sec:results-logit-lens}

\begin{figure*}[t]
    \centering
    \includegraphics[width=0.9\textwidth]{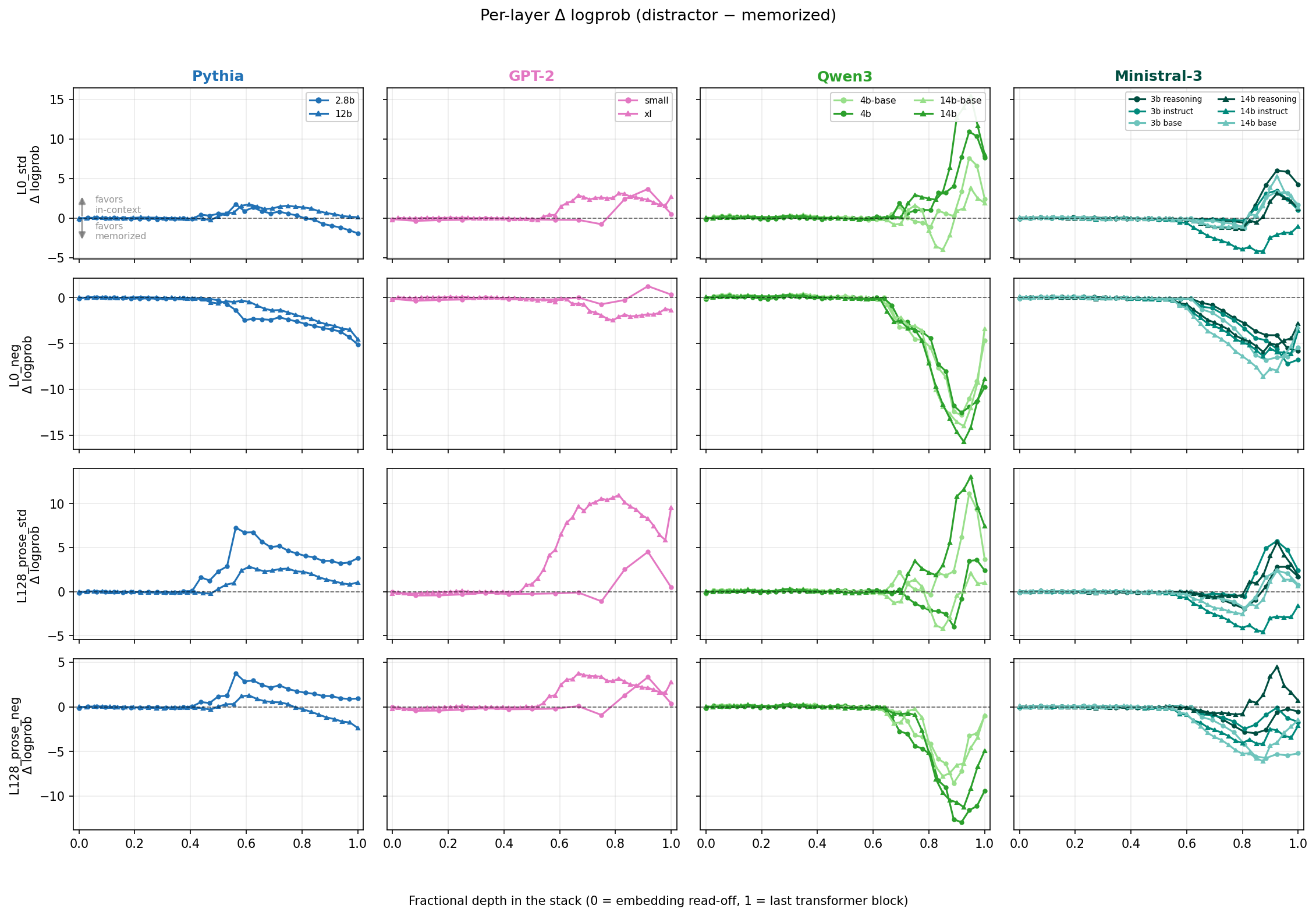}
    \caption[Cross-family logit lens on World Capital task]{Logit lens on the world-capitals dataset, indicating at each layer if a model leans towards the in-context or the memorized answer by decoding hidden states into vocabulary space.
        Rows: four matched conditions (no filler vs.\ $L{=}128$ prose;
        standard vs.\ negated conflict).}
    \label{fig:logit-lens-family}
\end{figure*}

Figure~\ref{fig:logit-lens-family} tracks the separation between memorized
and in-context answers across model families using the logit-lens protocol from
\S\ref{sec:methods-logit-lens}.
Across all model families, $\Delta_\ell$ separates decisively only in the second half of the network. This aligns with observations by \citet{meng2022locating}, who showed via causal tracing that while subject representations are processed in middle MLP layers, the factual prediction is formed and transferred to the residual stream primarily in later layers.

%% file: appendix.tex
\section{Baseline Factual Recall Without Conflict}
\label{sec:appendix-clean-factual-recall}

\begin{figure*}[t]
  \centering
  \includegraphics[width=\textwidth]{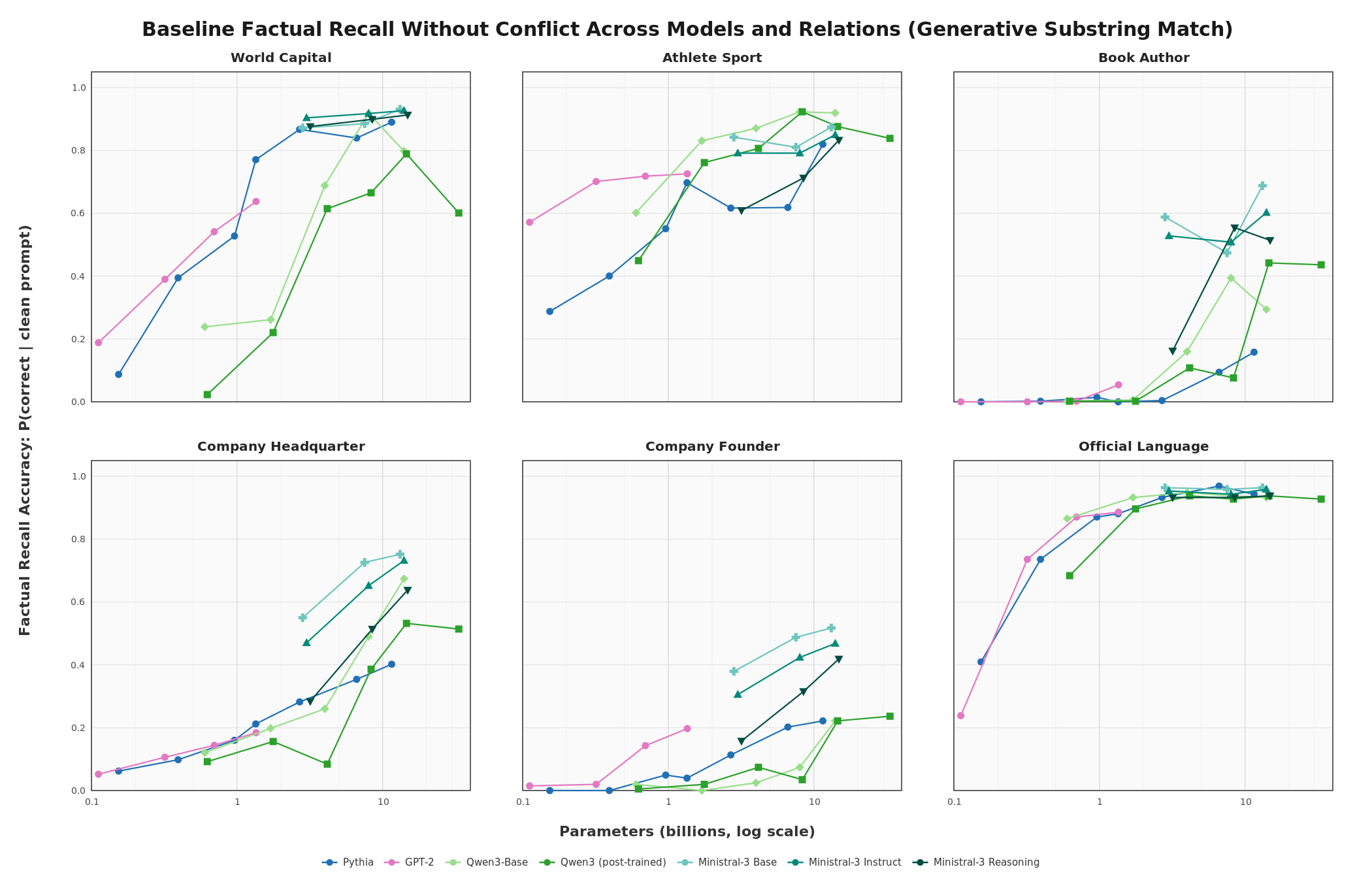}
  \caption{Baseline factual recall accuracy without conflict ($P(\text{correct} \mid \text{clean prompt})$; generative substring match) across model size for each ParaConflict relation category.}
  \label{fig:clean-factual-recall}
\end{figure*}

\begin{figure*}[t]
  \centering
  \includegraphics[width=\textwidth]{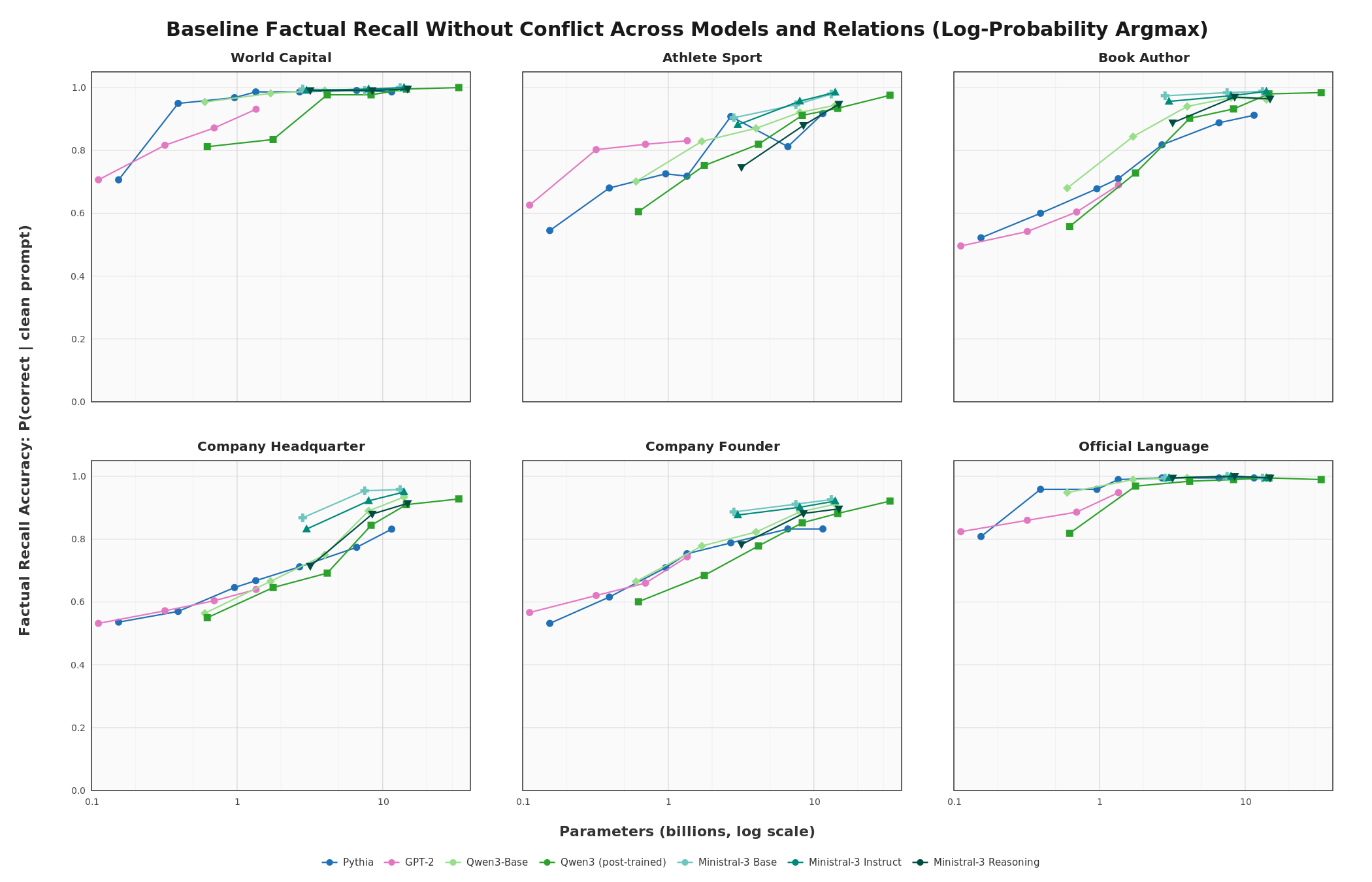}
  \caption{Baseline factual recall accuracy without conflict ($P(\text{correct} \mid \text{clean prompt})$; log-probability argmax) across model size for each ParaConflict relation category.}
  \label{fig:clean-factual-recall-lp}
\end{figure*}

This appendix supplements \S\ref{sec:results-model-size}.
To determine whether low $P(\text{mem})$ under substitution conflict is driven by context-following preference or by lack of parametric knowledge, we evaluate zero-shot factual recall without conflict ($n{=}2146$ prompts per model across all six categories) using the bare continuation clean prompt from the ParaConflict benchmark.
Figure~\ref{fig:clean-factual-recall} visualizes the scaling behavior across parameter counts for each relation category under generative substring evaluation, while Figure~\ref{fig:clean-factual-recall-lp} visualizes the corresponding accuracy under the log-probability classifier ($\arg\max_{y \in \{c_{\text{mem}},\, c_{\text{ctx}}\}} \log P(y \mid \text{clean prompt})$).

Notably, although small models have a low or near-zero generative factual recall rate on some datasets (such as Book Author or Company Founder; Figure~\ref{fig:clean-factual-recall}), Figure~\ref{fig:clean-factual-recall-lp} shows that they still prefer the gold answer over a random distractor when the distractor does not appear in the prompt (accuracies remain $\ge 50\%$ across all models and categories). This indicates that the models' parametric representations favor the correct entity over an unprompted distractor.

\section{Context-length: all fillers}
\label{sec:appendix-exp22-full}

\begin{figure*}[t]
  \centering
  \includegraphics[width=\textwidth]{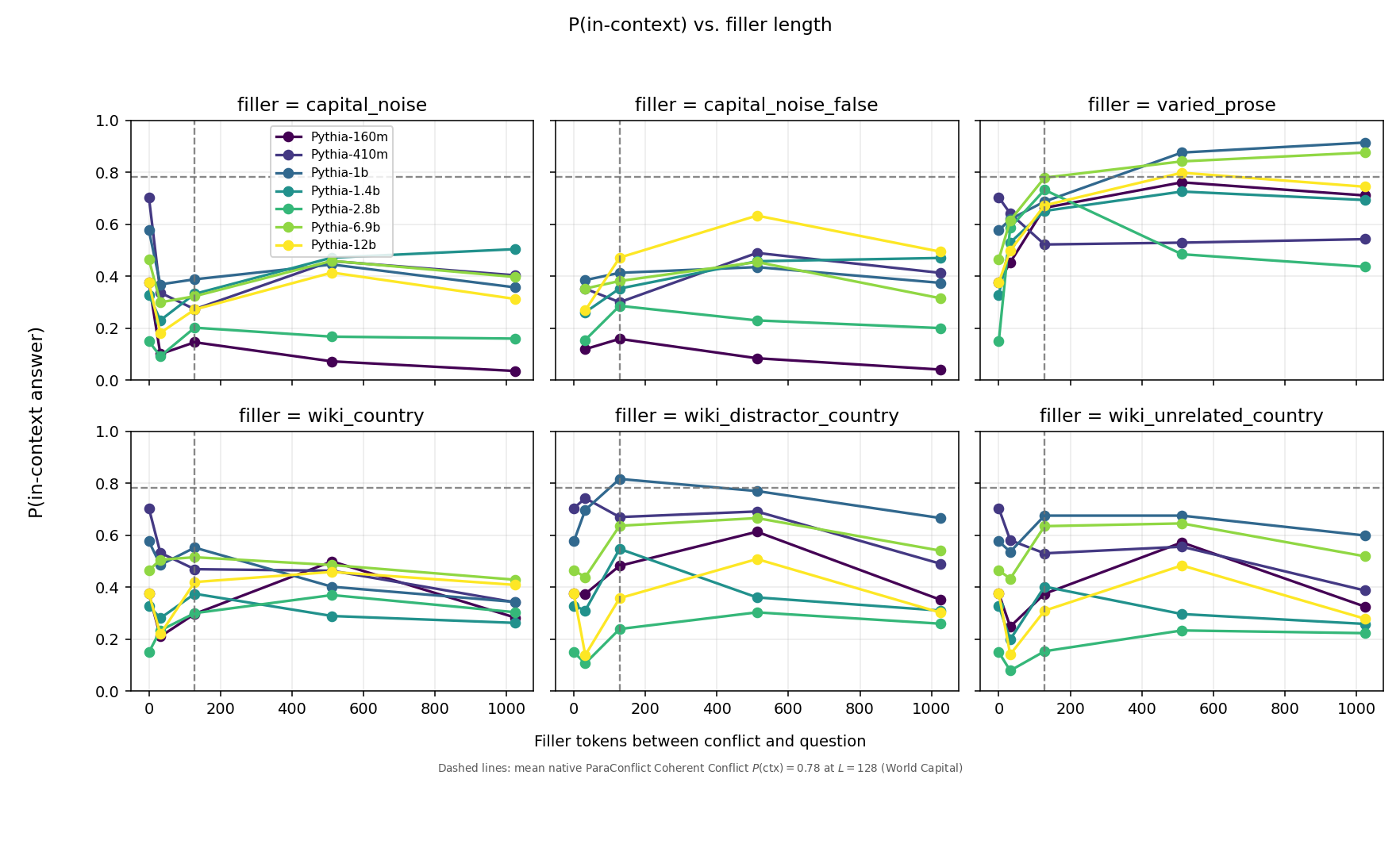}
  \caption{$P(\text{in-context})$ vs.\ filler length for every
    filler type and Pythia size.}
  \label{fig:exp22-gen-length-full}
\end{figure*}

This appendix supplements Figure~\ref{fig:exp22-gen-length} in
\S\ref{sec:results-context-sweep}.
Figure~\ref{fig:exp22-gen-length-full} shows the complete generative-classifier
grid over all six filler types at
$L \in \{0, 32, 128, 512, 1024\}$ (Pythia 160M--12B; $n{=}2000$ per cell). Surprisingly, the unrelated \texttt{prose} filler steer models towards the in-context answer, sometimes more than the coherent paragraph supporting the counterfactual claim (e.g for Pythia 1b and 6.7B with >512 filler tokens). Other fillers show mixed effects: a filler with text from the wikipedia page of the subject country, with the relevant capital redacted (\texttt{wiki\_country}) don't appear to influence the in-context-memory arbitration; one with the page of the distractor country (\texttt{wiki\_distractor\_country}) pushes the model towards ICL, but not significantly more than one from an unrelated country (\texttt{wiki\_unrelated\_country}).

\section{Cross-family generalization of unrelated prose (\texorpdfstring{$L=128$}{L=128})}
\label{sec:appendix-cross-family-prose}

\begin{figure*}[t]
  \centering
  \includegraphics[width=\textwidth]{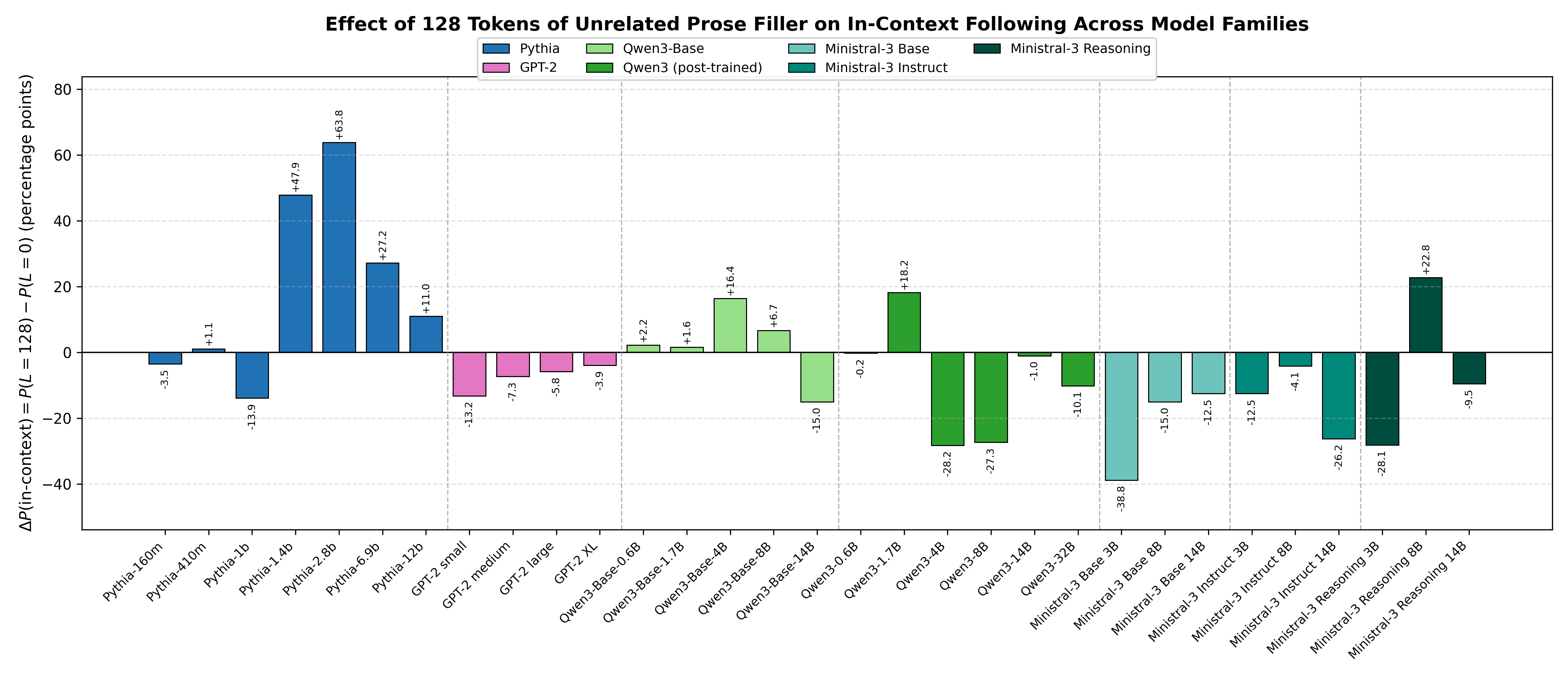}
  \caption{Cross-family effect of inserting $L{=}128$ tokens of unrelated biology prose between the conflict statement and the question across all 31 models ($n{=}2{,}000$ World Capital prompts, log-probability classifier). $\Delta P(\text{in-context}) = P(\text{in-context}\mid L{=}128) - P(\text{in-context}\mid L{=}0)$. Colors match Figure~\ref{fig:context_memory_score_gen_all} for each model family.}
  \label{fig:cross-family-prose-l128-delta}
\end{figure*}

To test whether the in-context boost from semantically unrelated prose observed on Pythia generalizes across other architectures, we evaluate all 31 models on $n{=}2{,}000$ World Capital queries comparing the standard zero-filler prompt ($L{=}0$) against an insertion of $L{=}128$ tokens of biology prose between the conflict statement and the question (tail position; identical to the setup in \S\ref{sec:results-context-sweep}).
Figure~\ref{fig:cross-family-prose-l128-delta} plots the percentage-point change $\Delta P(\text{in-context}) = P(\text{in-context} \mid L{=}128) - P(\text{in-context} \mid L{=}0)$ for each model.

Across models outside Pythia, the effect is inconsistent:
\begin{itemize}
  \item On Pythia, prose filler above 1B consistently increases in-context reliance (by up to $+63.8$\,pp on 2.8B). A similar boost appears on a few other models, such as Qwen3-Base-4B ($+16.4$\,pp), Qwen3-1.7B ($+18.2$\,pp), and Ministral-3 Reasoning-8B ($+22.8$\,pp).
  \item However, for many models across Qwen3 and Ministral-3, the effect moves strongly in the opposite direction: inserting unrelated prose reduces $P(\text{in-context})$ by up to $-28.2$\,pp (Qwen3-4B) and $-38.8$\,pp (Ministral-3 Base 3B), diluting the conflict statement and allowing the memorized prior to dominate.
\end{itemize}
Therefore, while coherent supporting context consistently raises $P(\text{in-context})$ across all 31 models (Finding~5, Table~\ref{tab:paraconflict-native-sub-vs-coh}), unrelated prose does not produce a consistent boost outside Pythia.

\section{Conflict position: front, middle, and tail}
\label{sec:appendix-position}

This appendix supplements the context-length analysis in
\S\ref{sec:results-context-sweep}.
The previous experiments holds the
conflict in \emph{tail} position
(\texttt{conflict\,|\,filler\,|\,question}).
Here we vary where the conflict statement sits relative to the filler
and question at $L \in \{128, 512\}$ tokens, using the paper
\texttt{qa} template, \texttt{prose} and \texttt{capital\_noise}
fillers, and the log-probability classifier ($n{=}2{,}000$ prompts per
cell; Pythia 160M--6.9B).
The three placements are:
\begin{itemize}
  \item \textbf{tail} --- \texttt{conflict\,|\,filler\,|\,question}
        (default elsewhere in the paper);
  \item \textbf{front} --- \texttt{filler\,|\,conflict\,|\,question}
        (conflict immediately before the question);
  \item \textbf{middle} --- \texttt{½filler\,|\,conflict\,|\,½filler\,\\ |\,question}
        (conflict embedded at the filler midpoint).
\end{itemize}

\begin{figure}[t]
  \centering
  \includegraphics[width=\columnwidth]{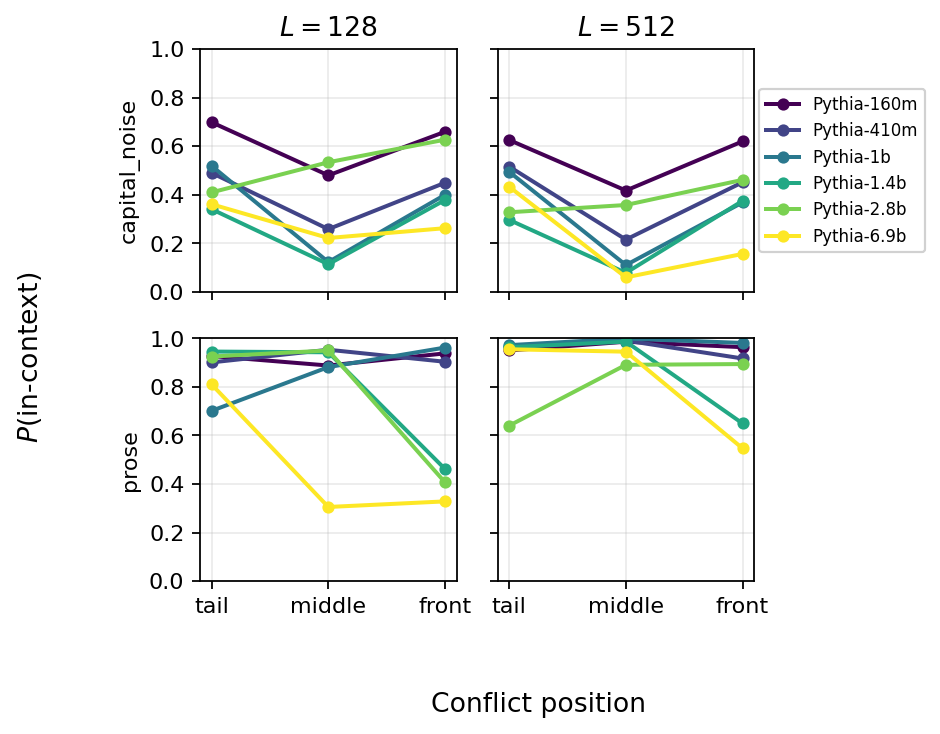}
  \caption{Effect of the position of the conflict in the prompt.}
  \label{fig:exp02-position}
\end{figure}

Figure~\ref{fig:exp02-position} shows $P(\text{in-context})$ across
all three positions.
With \texttt{capital\_noise}, placing the conflict in \texttt{middle}
is typically the worst cell for in-context binding---a
lost-in-the-middle-like pattern \cite{liu-etal-2024-lost}---with
competing capital statements on both sides (e.g.\ Pythia-1.4B at
$L{=}512$: 7.8\,\% vs.\ 29.6\,\% at \texttt{tail}).

\texttt{Prose} behaves differently: inserting filler \emph{between}
conflict and question (\texttt{tail}) steers models toward
in-context at medium-to-large scale, whereas placing the conflict
immediately before the question (\texttt{front}) often restores
memorization (Pythia-2.8B at $L{=}128$: 92.5\,\% $\to$ 40.7\,\%).
This is hard to reconcile with a simple induction-head recency story
\cite{olsson2022context}, which would predict stronger in-context use
when the conflict is last before the query.

\section{Post-training effects by model size}
\label{sec:appendix-posttraining}

This appendix supplements Table~\ref{tab:relations-posttraining} in
\S\ref{sec:results-model-size}.
Table~\ref{tab:relations-posttraining-by-size} reports
$\Delta P(\text{mem})$ for each size-matched post-training contrast
separately, without averaging across model scale.
The Qwen3-8B vs.\ Qwen3-Base-8B flip on book authors ($-59.8$\,pp) and
the Ministral-3-8B Instruct gain on the same relation ($+61.6$\,pp)
illustrate that pooled post-training effects can hide sharp
size-specific reversals.

\input{tables/relations_posttraining_by_size.tex}

\section{Model size vs.\ memorization by family}
\label{sec:appendix-size-mem-family}

This appendix supplements Table~\ref{tab:relations-size-mem} in
\S\ref{sec:results-model-size}.
Table~\ref{tab:relations-size-mem-by-family} shows the
size--memorization Spearman correlation separately within each model
family.
The all-model trend in Table~\ref{tab:relations-size-mem} masks
family-specific structure: Qwen3-Base is uniformly positive
($\bar{\rho} = +0.85$), whereas Pythia inverts on company headquarters
($\rho = -0.56$) and GPT-2 has $P(\text{mem}) = 0$ on three relations at
every size (undefined $\rho$ because on these tasks it never predicts the memorized response).

\input{tables/relations_size_mem_by_family.tex}

\section{Effect of Coherent Conflict}
\label{sec:appendix-native-sub-vs-coh}

Table~\ref{tab:paraconflict-native-sub-vs-coh} compares
\texttt{Substitution Conflict} and \texttt{Coherent Conflict} prompts on
all six ParaConflict relation categories ($n{=}2146$ per model;
generative classifier).  Replacing the minimal substitution
template with the dataset's fluent coherent passage raises
$P(\text{in-context})$ on every evaluated model
($\Delta > 0$ for all 31; median $\Delta = +0.19$;
mean $\Delta = +0.21$; range $+0.01$--$+0.57$).

\input{tables/paraconflict_native_sub_vs_coh.tex}

\section{Correlation Between Generative and Log-Probability Classifiers}
\label{sec:appendix-generative-logprob-correlation}
The two classifiers' $P(\mathrm{mem})$ estimates are very tightly correlated (Pearson $r \approx 0.98$); $P(\mathrm{ctx})$ less so ($r \approx 0.73$), mainly because greedy decoding sometimes emits a third answer that the log-probability classifier cannot assign. With filler tokens in the prompt, the correlation tends to decrease more for $P(\mathrm{ctx})$ due to the increased likelihood of generating a third answer.

\section{Country frequency and memorization: full results}
\label{sec:appendix-country-freq}

This appendix supplements \S\ref{sec:results-country-freq}.

\begin{figure*}[t]
  \centering
  \includegraphics[width=\textwidth]{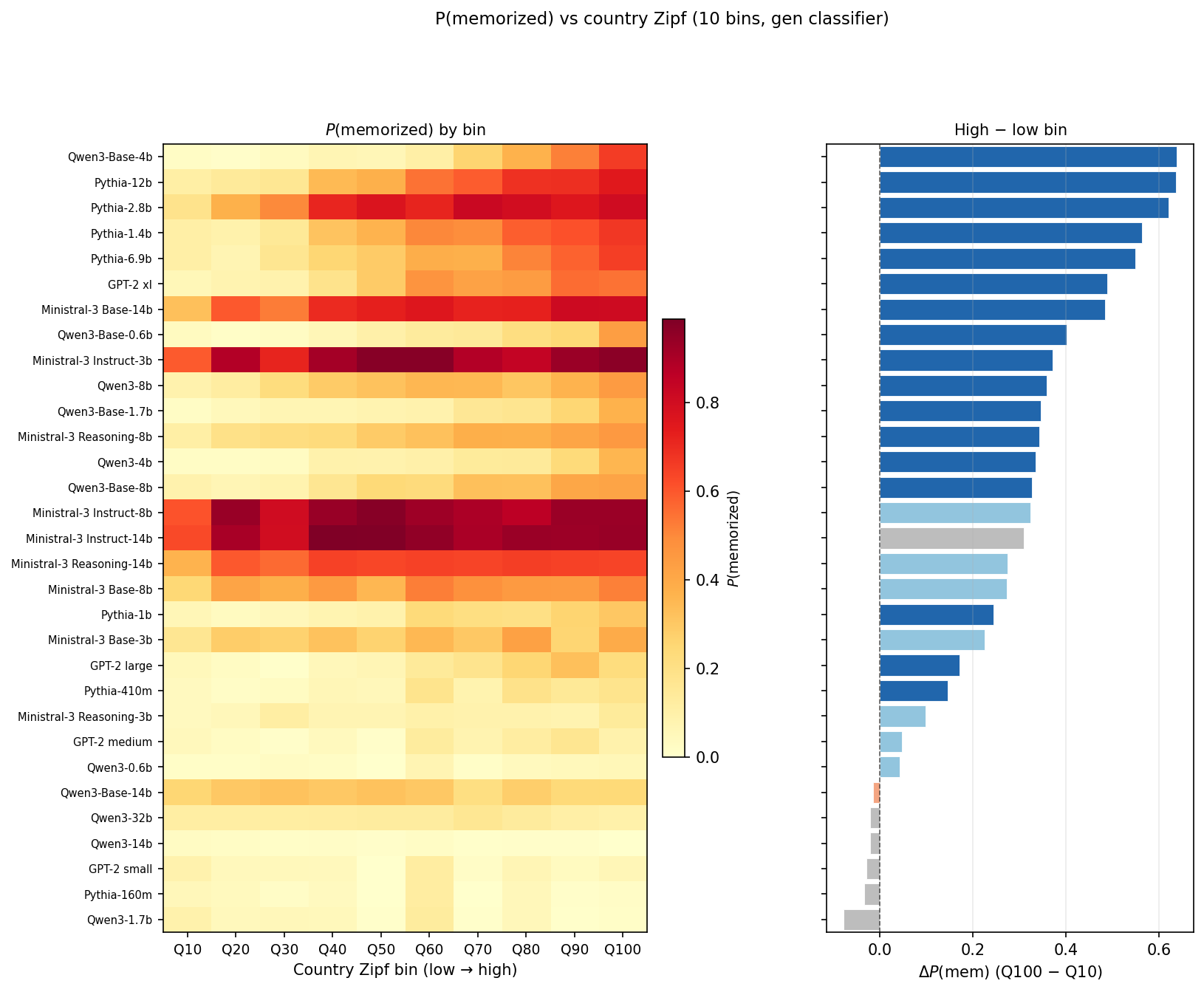}
  \caption{Country-frequency memorization across model families
    (generative classifier).  \textbf{Left:} $P(\text{memorized})$ in
    ten country-Zipf percentile bins (Q10 = rarest countries;
    Q100 = most frequent).  \textbf{Right:} $\Delta P(\text{memorized})$
    between Q100 and Q10.  Rows share a common $y$-axis and are sorted by
    $\Delta$ (ascending); bar color encodes the verdict in
    Table~\ref{tab:country-freq-mem} (blue = supports the Yu et al.\
    trend, red = inverts or weakly inverts, gray = flat).}
  \label{fig:country-freq-mem}
\end{figure*}

Table~\ref{tab:country-freq-mem}
lists all 31 models and associated statistics; Figure~\ref{fig:country-freq-mem} visualizes
the same data.

\input{tables/country_freq_mem_hypothesis.tex}

\section{ParaConflict entity word frequencies}
\label{sec:appendix-paraconflict-freq}

This appendix supplements the training-frequency proxy in
\S\ref{sec:methods-training-freq}.
For each of the six ParaConflict relation categories we collect unique
\emph{subjects} (queried entities) and \emph{memorized answers} (gold
objects from the \texttt{Answer} field), then score each
string with the \texttt{wordfreq} Zipf proxy (leading articles stripped;
same normalization as the country-frequency analysis).
Table~\ref{tab:paraconflict-freq-summary} reports the mean Zipf score
over unique entities in each role.

\input{tables/paraconflict_entity_frequency.tex}

\section{Alternative question template and cross-family phrasing sensitivity}
\label{sec:appendix-alt-scaffolds}

\begin{figure*}[t]
  \centering
  \includegraphics[width=\textwidth]{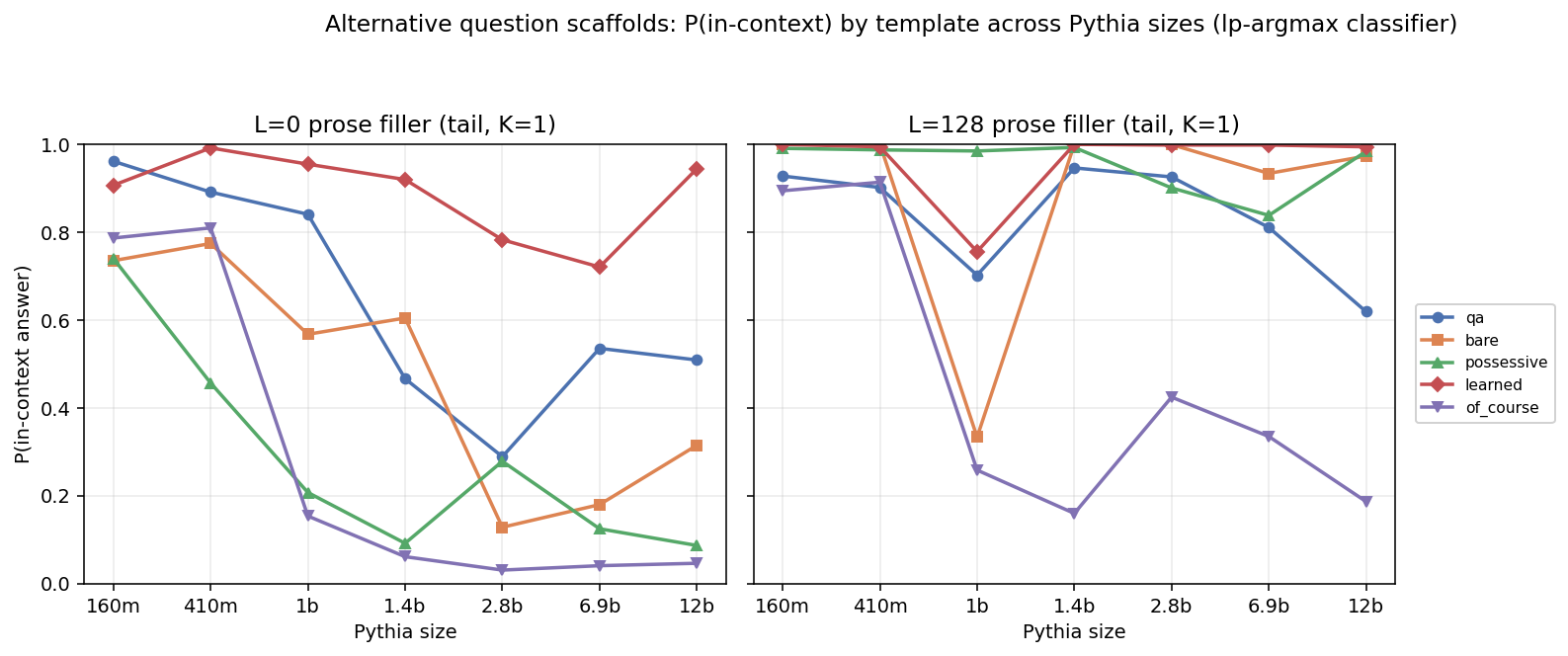}
  \caption{$P(\text{in-context})$ increasingly depends on the question template with Pythia model size.}
  \label{fig:exp12-template-comparison}
\end{figure*}

This appendix supplements the template ablation described in
\S\ref{sec:methods-template-ablations}.
First, to evaluate interaction with context length, we evaluate all Pythia sizes at $L \in \{0, 128\}$ with standard conflict, prose filler in tail position ($K{=}1$), and the log-probability classifier ($n{=}2{,}000$).
Figure~\ref{fig:exp12-template-comparison} shows that framing tokens
can swing $P(\text{in-context})$ by $\geq 80$ percentage points at fixed
conflict and filler.
\texttt{learned} saturates near 100\,\% in-context at $L{=}128$ for
most sizes, whereas \texttt{of\_course} suppresses in-context to
$\leq$20\,\% on Pythia $\geq$1.4B even with 128-token prose filler.
\texttt{possessive} strongly pushes the models to low $P(\text{in-context})$ without a filler but has the opposite effect with the 128-token filler.

\begin{figure*}[t]
  \centering
  \includegraphics[width=\textwidth]{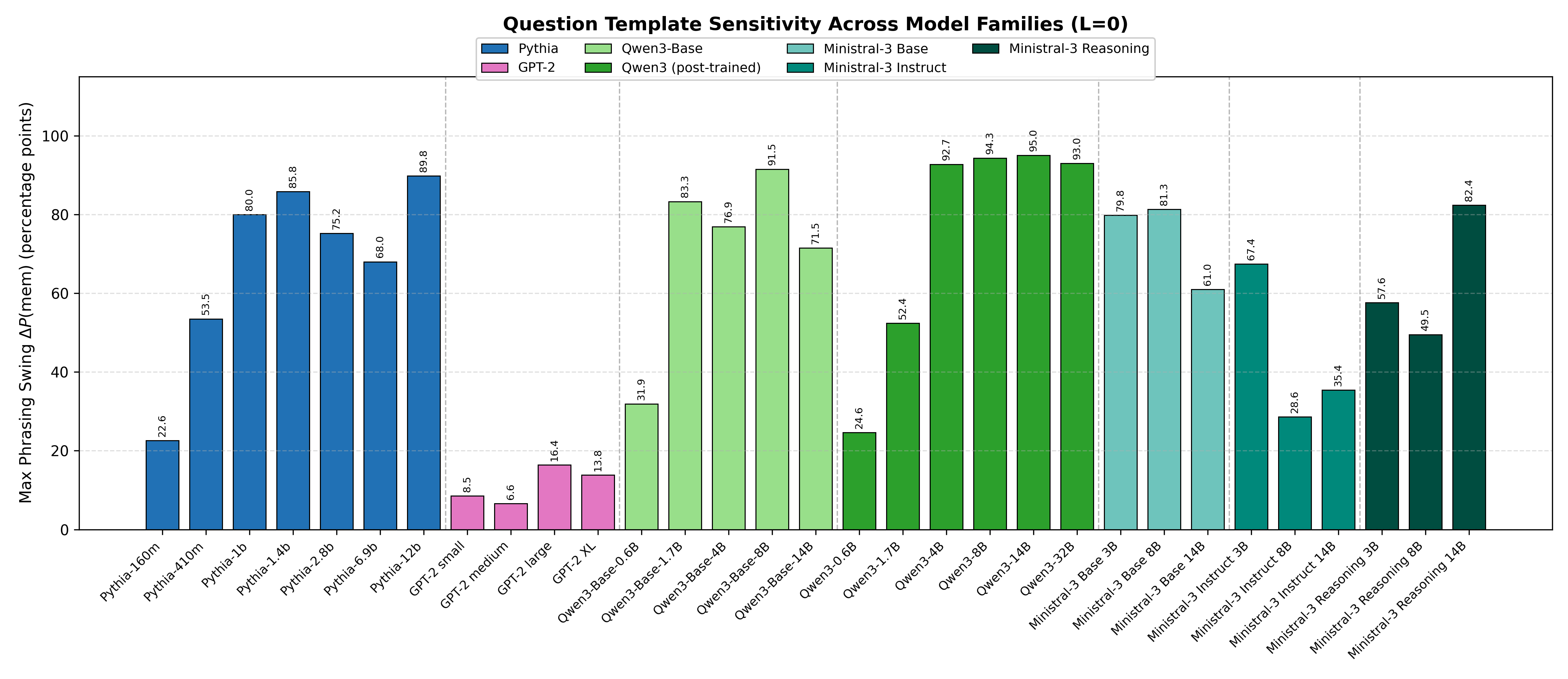}
  \caption{Maximum phrasing swing $\Delta_{\max} = \max P(\text{mem}) - \min P(\text{mem})$ across five question templates for all 31 models at $L{=}0$ ($n{=}2{,}000$).}
  \label{fig:cross-family-phrasing-swings}
\end{figure*}

\paragraph{Cross-family generalization across all 31 models.}
To test whether phrasing sensitivity is an artifact of the Pythia suite, we extend the five-template evaluation (\texttt{qa}, \texttt{bare}, \texttt{possessive}, \texttt{learned}, \texttt{of\_course}) across all 31 models in our study at $L{=}0$ ($n{=}2{,}000$ prompts per cell; 155 evaluations total).
Figure~\ref{fig:cross-family-phrasing-swings} reports the resulting maximum template swings $\Delta_{\max} = \max P(\text{mem}) - \min P(\text{mem})$ across the five templates.

While older, smaller architectures (GPT-2) display modest swings (6.6 to 16.4\,pp), modern pre-trained and post-trained models exhibit massive phrasing sensitivity across every family:
\begin{itemize}
  \item \textbf{Pythia}: swings range from 22.6\,pp (160M) to 89.8\,pp (12B), growing with scale.
  \item \textbf{Qwen3-Base}: swings reach 83.3\,pp (1.7B), 91.5\,pp (8B), and 71.5\,pp (14B).
  \item \textbf{Qwen3 (post-trained)}: swings reach 92.7\,pp (4B), 94.3\,pp (8B), 95.0\,pp (14B), and 93.0\,pp (32B). Models that appear almost completely immune to memorized knowledge under the standard \texttt{qa} scaffold ($0.0\%$ for 14B, $2.1\%$ for 32B) invert entirely under \texttt{of\_course} (jumping to $95.0\%$ and $95.2\%$ memorization, respectively) and under \texttt{bare} ($71.0\%$ for 14B).
  \item \textbf{Ministral-3}: swings reach 81.3\,pp in Base (8B), 67.4\,pp in Instruct (3B), and 82.4\,pp in Reasoning (14B).
\end{itemize}
These results confirm that prompt phrasing sensitivity is an architecture-general phenomenon and that measuring conflict arbitration from any single question scaffold can yield highly misleading conclusions about model-level memory resistance.

\section{Negated conflict statement}
\label{sec:appendix-negated-conflict}

This appendix supplements the conflict-template ablation in
\S\ref{sec:methods-template-ablations}.
We replace \texttt{...is \{distractor\}.} with
\texttt{...is not \{distractor\}.} under the paper \texttt{qa} template,
tail position ($K{=}1$), the log-probability classifier, and
$n{=}2{,}000$ prompts at $L \in \{0, 128\}$ with prose or
\texttt{capital\_noise} filler.
Table~\ref{tab:negated-conflict} reports $P(\text{in-context})$ under
standard vs.\ negated conflict; cells show standard $\to$ negated
($\Delta$).
Negative $\Delta$ indicates a shift toward the memorized capital.
At $L{=}0$ the two fillers agree (no intervening tokens); negation
has little effect on Pythia-160M but grows with scale, reaching
$\Delta = -42$ for Pythia-12B.
At $L{=}128$, both fillers weaken this effect, \texttt{capital\_noise} more than \texttt{prose}.

\begin{table*}[t]
  \centering
  \footnotesize
  \setlength{\tabcolsep}{4pt}
  \caption{Negated vs.\ standard conflict (log-probability classifier).
    Each cell gives $P(\text{in-context}\mid\text{standard})
      \to P(\text{in-context}\mid\text{negated})$ ($\Delta$).}
  \label{tab:negated-conflict}
  \begin{tabular*}{\textwidth}{@{\extracolsep{\fill}}l ccc@{}}
    \toprule
    Size         & $L{=}0$                            & $L{=}128$ (prose)                  & $L{=}128$ (\texttt{capital\_noise}) \\
    \midrule
    160m         & 0.96 $\to$ 0.97 ($+$0.01)          & 0.93 $\to$ 0.94 ($+$0.02)          & 0.70 $\to$ 0.76 ($+$0.06)           \\
    410m         & 0.89 $\to$ 0.81 ($-$0.08)          & 0.90 $\to$ 0.84 ($-$0.06)          & 0.49 $\to$ 0.46 ($-$0.03)           \\
    1b           & 0.84 $\to$ 0.68 ($-$0.16)          & 0.70 $\to$ 0.51 ($-$0.20)          & 0.52 $\to$ 0.40 ($-$0.12)           \\
    1.4b         & 0.47 $\to$ 0.27 ($-$0.20)          & 0.95 $\to$ 0.83 ($-$0.11)          & 0.34 $\to$ 0.26 ($-$0.08)           \\
    2.8b         & 0.29 $\to$ 0.09 ($-$0.20)          & 0.93 $\to$ 0.64 ($-$0.29)          & 0.41 $\to$ 0.20 ($-$0.21)           \\
    6.9b         & 0.54 $\to$ 0.23 ($-$0.30)          & 0.81 $\to$ 0.52 ($-$0.29)          & 0.36 $\to$ 0.24 ($-$0.12)           \\
    \textbf{12b} & \textbf{0.51 $\to$ 0.09 ($-$0.42)} & \textbf{0.62 $\to$ 0.26 ($-$0.36)} & \textbf{0.31 $\to$ 0.19 ($-$0.12)}  \\
    \bottomrule
  \end{tabular*}
\end{table*}

\section{Conflict repetition}
\label{sec:appendix-repetition-sweep}

This appendix supplements the conflict-copy ablation in
\S\ref{sec:methods-template-ablations}.
With no intervening filler ($L{=}0$), we repeat the standard conflict
statement $K \in \{1, 2, 4, 8\}$ times before the \texttt{qa}
question, holding all else fixed (log-probability classifier,
$n{=}2{,}000$ prompts, Pythia suite).

\begin{figure}[t]
  \centering
  \includegraphics[width=\columnwidth]{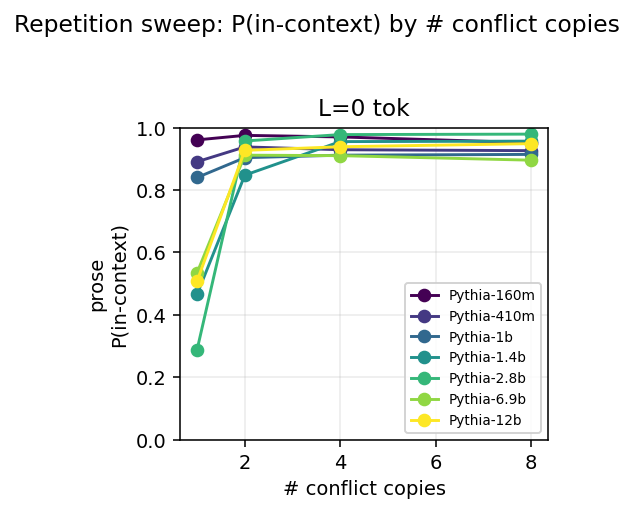}
  \caption{Conflict repetition at $L{=}0$ (log-probability classifier).
    $P(\text{in-context})$ vs.\ Pythia size for
    $K \in \{1, 2, 4, 8\}$ redundant conflict copies.}
  \label{fig:exp07-repetition}
\end{figure}

Figure~\ref{fig:exp07-repetition} shows that a single extra copy ($K{=}2$) pushes every size
to 85--97\,\% in-context; $K{=}4$ and $K{=}8$ plateau near saturation.
Repeated same-template assertions therefore reinforce in-context binding
even without recency from filler tokens. This is a much stronger effect than
\texttt{capital\_noise} filler.

\section{Model Checkpoints and Inference Interface}
\label{sec:appendix-inference-specs}

Table~\ref{tab:model-inference-specs} lists all 31 evaluated checkpoints across the Pythia, GPT-2, Qwen3, and Ministral-3 families. Evaluations use native \texttt{bfloat16} (Qwen3, Ministral-3), \texttt{float16} (Pythia-6.9B, 12B), and \texttt{float32} (smaller Pythia, GPT-2).

\input{tables/model_inference_specs.tex}

\paragraph{Prompt formatting}
All models are evaluated on identical raw completion prompts (e.g., \texttt{"The capital of Poland is London. Q: What is the capital of Poland? A:"}) without chat templates (e.g., ChatML \texttt{<|im\_start|>} or \texttt{[INST]}). As shown in \S\ref{sec:results-model-size}, prompt phrasing alone swings $P(\text{mem})$ by up to 95\,pp; wrapping instruct/reasoning models in chat templates while feeding raw text to base models would conflate post-training effects with prompt framing. Raw evaluation maintains token-level input parity across stages and architectures.

\paragraph{Reasoning models and generation limits.}
Generative evaluation uses deterministic greedy decoding ($T{=}0$, $\text{max\_new\_tokens}{=}10$) with substring matching against gold aliases. We acknowledge that this uniform evaluation design choice does not equally test reasoning models under the conditions for which they are designed, such as extended generation budgets or multi-step thinking processes. Rather than introducing model-specific generation limits or specialized reasoning scaffolds, we maintain identical decoding constraints across all 31 base, instruction, and reasoning models to preserve controlled comparability, even though this uniform 10-token regime restricts reasoning models to direct short-horizon completion.

%% file: tables/relations_posttraining_by_size.tex
\begin{table*}[t]
  \centering
  \footnotesize
  \caption{Per-model effect of post-training on $P(\text{mem})$
    (percentage-point change, post-trained minus Base) for size-matched
    models on ParaConflict \texttt{Substitution Conflict} prompts. No significant trend change is observed across model size for either Qwen3 or Ministral-3.}
  \label{tab:relations-posttraining-by-size}
  \adjustbox{max width=\textwidth}{%
  \setlength{\tabcolsep}{3.2pt}%
  \begin{tabular}{@{}l rrrrr rrr rrr@{}}
    \toprule
     & \multicolumn{5}{c}{Qwen3 vs.\ Base}
     & \multicolumn{3}{c}{Ministral-3 Instruct vs.\ Base}
     & \multicolumn{3}{c}{Ministral-3 Reasoning vs.\ Base}                                         \\
    \cmidrule(lr){2-6}\cmidrule(lr){7-9}\cmidrule(lr){10-12}
    Category
     & 0.6B                                              & 1.7B    & 4B      & 8B      & 14B
     & 3B                                                & 8B      & 14B
     & 3B                                                & 8B      & 14B                         \\
    \midrule
    World Capital
     & $-$11.0                                           & $-$8.8  & $-$9.0  & $+$5.2  & $-$26.2
     & $+$56.5                                           & $+$45.3 & $+$22.3
     & $-$21.9                                           & $-$12.8 & $-$6.8                      \\
    Athlete Sport
     & $-$0.2                                            & $-$3.2  & $+$1.1  & $-$13.7 & $-$43.0
     & $+$5.1                                            & $+$7.0  & $+$9.4
     & $-$3.0                                            & $-$13.0 & $+$7.0                      \\
    Book Author
     & $+$0.0                                            & $+$3.4  & $-$7.8  & $-$59.8 & $+$7.2
     & $-$3.4                                            & $+$61.6 & $+$55.6
     & $-$11.2                                           & $+$32.2 & $-$7.4                      \\
    Company Headquarter
     & $+$0.2                                            & $+$1.0  & $+$0.4  & $-$6.0  & $+$4.8
     & $+$8.0                                            & $+$9.2  & $+$19.2
     & $+$0.2                                            & $-$1.8  & $-$5.6                      \\
    Company Founder
     & $+$0.0                                            & $-$6.4  & $-$9.4  & $-$10.3 & $-$29.6
     & $+$1.0                                            & $+$27.6 & $+$22.7
     & $-$2.5                                            & $+$14.8 & $-$3.9                      \\
    Official Language
     & $-$62.2                                           & $-$29.0 & $-$3.1  & $-$55.4 & $-$9.3
     & $+$10.9                                           & $+$15.0 & $+$4.1
     & $+$8.8                                            & $-$17.6 & $-$45.6                     \\
    \midrule
    Mean (relations)
     & $-$12.2                                           & $-$7.2  & $-$4.6  & $-$23.3 & $-$16.0
     & $+$13.0                                           & $+$27.6 & $+$22.2
     & $-$4.9                                            & $+$0.3  & $-$10.4                     \\
    \bottomrule
  \end{tabular}%
  }
\end{table*}

%% file: tables/relations_size_mem_by_family.tex
\begin{table*}[t]
  \centering
  \footnotesize
  \caption{Spearman $\rho$ between parameter count and $P(\text{mem})$ by
    model family.}
  \label{tab:relations-size-mem-by-family}
  \adjustbox{max width=\textwidth}{%
  \setlength{\tabcolsep}{4.5pt}%
  \begin{tabular}{@{}l rrrrrr rr@{}}
    \toprule
    Family ($n_{\text{models}}$)
     & World Capital & Athlete Sport & Book Author & Co.\ HQ & Co.\ Founder & Official Lang.
     & $\bar{P}(\text{mem})$ & $\bar{\rho}$ \\
    \midrule
    Pythia (7)
     & $+$0.82               & $+$0.68      & $+$0.63 & $-$0.56 & $+$0.76 & $+$0.68
     & 0.18                  & $+$0.50                                              \\
    GPT-2 (4)
     & $+$1.00               & $-$0.20      & ---     & ---     & ---     & $-$0.20
     & 0.06                  & $+$0.20                                              \\
    Qwen3-Base (5)
     & $+$0.90               & $+$0.90      & $+$0.70 & $+$1.00 & $+$0.90 & $+$0.70
     & 0.31                  & $+$0.85                                              \\
    Qwen3 (6)
     & $+$0.14               & $+$0.94      & $+$0.77 & $+$0.94 & $+$0.93 & $+$0.60
     & 0.21                  & $+$0.72                                              \\
    \shortstack{Ministral-3\\Base} (3)
     & $+$1.00               & $+$0.50      & $+$0.50 & $+$1.00 & $+$0.50 & $-$0.50
     & 0.27                  & $+$0.50                                              \\
    \shortstack{Ministral-3\\Instruct} (3)
     & $+$1.00               & $+$0.50      & $+$1.00 & $+$1.00 & $+$0.50 & $-$0.50
     & 0.48                  & $+$0.58                                              \\
    \shortstack{Ministral-3\\Reasoning} (3)
     & $+$1.00               & $+$1.00      & $+$0.50 & $+$1.00 & $+$0.50 & $-$1.00
     & 0.22                  & $+$0.50                                              \\
    \bottomrule
  \end{tabular}%
  }
\end{table*}

%% file: tables/paraconflict_native_sub_vs_coh.tex
\begin{table}[t]
      \centering
      \small
      \setlength{\tabcolsep}{0pt}
      \caption{Native ParaConflict: $P(\text{in-context})$ under Substitution vs.\ Coherent Conflict (generative classifier; six categories; $n{=}2146$ per model). $\Delta = P(\text{ctx}\mid\text{Coh}) - P(\text{ctx}\mid\text{Sub})$.}
      \label{tab:paraconflict-native-sub-vs-coh}
      \begin{tabular*}{\columnwidth}{@{\extracolsep{\fill}}lccc@{}}
            \toprule
            Model                     & $P(\text{ctx}\mid\text{Sub})$ & $P(\text{ctx}\mid\text{Coh})$ & $\Delta$ \\
            \midrule
            \multicolumn{4}{l}{\textbf{Pythia}}                                                                  \\
            Pythia-160m               & 0.71                          & 0.82                          & +0.11    \\
            Pythia-410m               & 0.86                          & 0.92                          & +0.06    \\
            Pythia-1b                 & 0.76                          & 0.92                          & +0.16    \\
            Pythia-1.4b               & 0.85                          & 0.92                          & +0.08    \\
            Pythia-2.8b               & 0.76                          & 0.93                          & +0.17    \\
            Pythia-6.9b               & 0.70                          & 0.94                          & +0.24    \\
            Pythia-12b                & 0.81                          & 0.92                          & +0.12    \\
            \addlinespace
            \multicolumn{4}{l}{\textbf{GPT-2}}                                                                   \\
            GPT-2~small               & 0.90                          & 0.92                          & +0.02    \\
            GPT-2~medium              & 0.94                          & 0.95                          & +0.01    \\
            GPT-2~large               & 0.88                          & 0.94                          & +0.05    \\
            GPT-2~xl                  & 0.93                          & 0.94                          & +0.01    \\
            \addlinespace
            \multicolumn{4}{l}{\textbf{Qwen3}}                                                                   \\
            Qwen3-0.6B                & 0.76                          & 0.97                          & +0.21    \\
            Qwen3-1.7B                & 0.80                          & 0.93                          & +0.14    \\
            Qwen3-4B                  & 0.64                          & 0.90                          & +0.27    \\
            Qwen3-8B                  & 0.71                          & 0.94                          & +0.23    \\
            Qwen3-14B                 & 0.49                          & 0.96                          & +0.47    \\
            Qwen3-32B                 & 0.48                          & 0.94                          & +0.47    \\
            \addlinespace
            \multicolumn{4}{l}{\textbf{Qwen3-Base}}                                                              \\
            Qwen3-Base-0.6B           & 0.89                          & 0.95                          & +0.07    \\
            Qwen3-Base-1.7B           & 0.77                          & 0.96                          & +0.18    \\
            Qwen3-Base-4B             & 0.61                          & 0.95                          & +0.34    \\
            Qwen3-Base-8B             & 0.39                          & 0.97                          & +0.57    \\
            Qwen3-Base-14B            & 0.40                          & 0.96                          & +0.56    \\
            \addlinespace
            \multicolumn{4}{l}{\textbf{Ministral-3}}                                                             \\
            Ministral-3~Instruct~3B   & 0.67                          & 0.71                          & +0.05    \\
            Ministral-3~Instruct~8B   & 0.44                          & 0.76                          & +0.32    \\
            Ministral-3~Instruct~14B  & 0.44                          & 0.75                          & +0.31    \\
            Ministral-3~Reasoning~3B  & 0.78                          & 0.97                          & +0.19    \\
            Ministral-3~Reasoning~8B  & 0.67                          & 0.97                          & +0.31    \\
            Ministral-3~Reasoning~14B & 0.81                          & 0.97                          & +0.16    \\
            Ministral-3~Base~3B       & 0.79                          & 0.97                          & +0.19    \\
            Ministral-3~Base~8B       & 0.76                          & 0.97                          & +0.22    \\
            Ministral-3~Base~14B      & 0.73                          & 0.97                          & +0.24    \\
            \bottomrule
      \end{tabular*}
\end{table}

%% file: tables/country_freq_mem_hypothesis.tex
\begin{table*}[t]
  \centering
  \scriptsize
  \setlength{\tabcolsep}{3pt}
  \caption{$P(\text{mem})$ by queried-country \texttt{wordfreq} Zipf percentile bin (10 equal-count bins; $n{\approx}4{,}730$ prompts per bin per model). $\rho$: Spearman correlation between per-country frequency and $P(\text{mem})$; $\Delta$: $P(\text{mem})$ in the highest minus lowest bin (percentage points). Verdicts follow \S\ref{sec:results-country-freq}: \emph{supports} if $\rho \geq 0.3$ and $\Delta \geq 5$\,pp.}
  \label{tab:country-freq-mem}
  \begin{tabular*}{\textwidth}{@{\extracolsep{\fill}}l rrrrrrrrrr rr l@{}}
    \toprule
    Model & Q10 & Q20 & Q30 & Q40 & Q50 & Q60 & Q70 & Q80 & Q90 & Q100 & $\rho$  & $\Delta$\,pp & Verdict \\
    \midrule
    GPT-2 small & 8.8 & 4.5 & 4.7 & 4.6 & 0.0 & 12.3 & 1.7 & 6.5 & 3.5 & 5.8 & +0.430 & -2.9 & flat \\
    GPT-2 medium & 4.4 & 2.6 & 1.1 & 4.1 & 1.1 & 12.9 & 8.0 & 12.1 & 17.0 & 9.2 & +0.553 & +4.7 & weak support \\
    GPT-2 large & 4.9 & 2.4 & 0.6 & 5.2 & 6.2 & 14.1 & 17.4 & 25.2 & 32.5 & 22.0 & +0.613 & +17.1 & \textbf{supports} \\
    GPT-2 XL & 5.8 & 7.8 & 8.6 & 18.5 & 29.3 & 46.9 & 42.4 & 44.0 & 55.8 & 54.7 & +0.642 & +48.9 & \textbf{supports} \\
    Ministral-3 Base-3b & 16.6 & 28.5 & 26.7 & 31.9 & 26.4 & 34.7 & 29.9 & 42.5 & 25.8 & 39.1 & +0.242  & +22.5 & weak support \\
    Ministral-3 Base-8b & 24.5 & 41.5 & 38.2 & 45.1 & 35.5 & 52.3 & 48.1 & 44.9 & 44.4 & 51.7 & +0.228  & +27.2 & weak support \\
    Ministral-3 Base-14b & 32.4 & 59.7 & 52.7 & 70.1 & 72.7 & 75.8 & 71.8 & 72.5 & 81.3 & 80.8 & +0.383 & +48.4 & \textbf{supports} \\
    Ministral-3 Instruct-3b & 59.1 & 88.3 & 71.4 & 91.1 & 97.2 & 97.1 & 88.2 & 84.1 & 93.3 & 96.3 & +0.366  & +37.2 & \textbf{supports} \\
    Ministral-3 Instruct-8b & 60.8 & 93.6 & 80.4 & 93.6 & 96.9 & 92.4 & 89.6 & 86.4 & 93.1 & 93.1 & +0.061  & +32.4 & weak support \\
    Ministral-3 Instruct-14b & 62.6 & 90.6 & 79.9 & 98.8 & 98.0 & 95.3 & 90.0 & 93.1 & 92.6 & 93.4 & -0.036 & +30.9 & flat \\
    Ministral-3 Reasoning-3b & 3.8 & 5.3 & 11.4 & 7.1 & 7.3 & 9.6 & 8.7 & 8.5 & 8.0 & 13.5 & +0.284  & +9.8 & weak support \\
    Ministral-3 Reasoning-8b & 11.2 & 19.4 & 21.9 & 23.1 & 29.2 & 32.3 & 37.9 & 37.5 & 41.1 & 45.4 & +0.521  & +34.2 & \textbf{supports} \\
    Ministral-3 Reasoning-14b & 36.4 & 59.5 & 56.1 & 64.7 & 63.6 & 64.6 & 64.0 & 65.4 & 64.5 & 63.9 & +0.201  & +27.5 & weak support \\
    Pythia-160m & 5.3 & 4.2 & 1.6 & 3.6 & 0.0 & 12.1 & 0.1 & 5.0 & 1.0 & 1.9 & +0.274  & -3.4 & flat \\
    Pythia-410m & 3.5 & 1.7 & 3.0 & 5.9 & 5.2 & 18.0 & 8.2 & 18.8 & 14.7 & 18.0 & +0.543  & +14.5 & \textbf{supports} \\
    Pythia-1b & 6.0 & 3.3 & 5.3 & 7.6 & 9.1 & 23.2 & 21.0 & 20.3 & 26.1 & 30.4 & +0.510  & +24.4 & \textbf{supports} \\
    Pythia-1.4b & 10.8 & 8.9 & 15.0 & 31.4 & 36.6 & 50.2 & 48.6 & 58.4 & 61.2 & 67.1 & +0.571  & +56.3 & \textbf{supports} \\
    Pythia-2.8b & 18.2 & 37.3 & 50.1 & 71.3 & 76.5 & 71.6 & 82.5 & 79.6 & 75.6 & 80.2 & +0.513  & +62.1 & \textbf{supports} \\
    Pythia-6.9b & 10.4 & 7.1 & 16.8 & 25.1 & 29.0 & 38.3 & 37.5 & 51.0 & 57.7 & 65.3 & +0.602  & +54.9 & \textbf{supports} \\
    Pythia-12b & 10.9 & 14.2 & 16.1 & 34.5 & 37.7 & 54.4 & 58.9 & 68.5 & 68.9 & 74.5 & +0.625  & +63.7 & \textbf{supports} \\
    Qwen3-0.6b & 1.2 & 1.4 & 2.4 & 2.0 & 0.1 & 7.2 & 1.5 & 4.1 & 4.7 & 5.4 & +0.451 & +4.2 & weak support \\
    Qwen3-1.7b & 9.1 & 4.5 & 5.0 & 4.6 & 0.6 & 13.1 & 0.5 & 5.3 & 0.8 & 1.3 & +0.288  & -7.8 & flat \\
    Qwen3-4b & 2.1 & 2.1 & 2.9 & 9.1 & 8.8 & 9.7 & 13.8 & 14.0 & 23.4 & 35.7 & +0.679  & +33.6 & \textbf{supports} \\
    Qwen3-8b & 8.7 & 12.3 & 22.1 & 29.0 & 31.7 & 35.3 & 34.8 & 30.8 & 36.6 & 44.6 & +0.436  & +35.9 & \textbf{supports} \\
    Qwen3-14b & 2.4 & 2.1 & 1.2 & 1.9 & 0.8 & 2.2 & 0.5 & 0.9 & 0.9 & 0.3 & +0.009  & -2.1 & flat \\
    Qwen3-32b & 11.4 & 11.4 & 11.7 & 12.1 & 12.6 & 12.6 & 15.9 & 13.3 & 10.8 & 9.3 & +0.073 & -2.0 & flat \\
    Qwen3-Base-0.6b & 3.1 & 1.5 & 2.6 & 5.8 & 9.8 & 13.3 & 14.5 & 21.5 & 24.4 & 43.3 & +0.716  & +40.2 & \textbf{supports} \\
    Qwen3-Base-1.7b & 2.0 & 5.0 & 6.9 & 6.9 & 8.1 & 8.7 & 15.7 & 17.0 & 25.2 & 36.7 & +0.661  & +34.7 & \textbf{supports} \\
    Qwen3-Base-4b & 2.0 & 1.1 & 3.3 & 6.9 & 5.9 & 10.5 & 26.0 & 36.8 & 52.1 & 65.9 & +0.678  & +63.9 & \textbf{supports} \\
    Qwen3-Base-8b & 8.6 & 6.2 & 8.4 & 16.5 & 24.2 & 23.1 & 32.7 & 32.1 & 40.6 & 41.3 & +0.581 &  +32.7 & \textbf{supports} \\
    Qwen3-Base-14b & 25.3 & 29.8 & 31.6 & 30.1 & 31.8 & 30.0 & 20.9 & 28.0 & 23.8 & 24.0 & -0.022 &  -1.4 & weak invert \\
    \bottomrule
  \end{tabular*}
\end{table*}

%% file: tables/paraconflict_entity_frequency.tex
\begin{table}[t]
  \centering
  \small
  \setlength{\tabcolsep}{4pt}
  \caption{Mean \texttt{wordfreq} Zipf per unique entity by ParaConflict category. Each cell averages Zipf over distinct subject or memorized-answer strings; leading \texttt{the} is stripped before scoring.}
  \label{tab:paraconflict-freq-summary}
  \begin{tabular}{@{}lrrrr@{}}
    \toprule
                        & \multicolumn{2}{c}{Subject} & \multicolumn{2}{c}{Memorized answer}                   \\
    \cmidrule(lr){2-3} \cmidrule(lr){4-5}
    Category            & $n$                         & $\bar{z}$                            & $n$ & $\bar{z}$ \\
    \midrule
    World~Capital       & 218                         & 3.58                                 & 260 & 2.67      \\
    Athlete~Sport       & 532                         & 2.93                                 & 33  & 4.09      \\
    Book~Author         & 500                         & 3.42                                 & 279 & 2.68      \\
    Company~Headquarter & 500                         & 3.02                                 & 242 & 3.37      \\
    Company~Founder     & 203                         & 2.93                                 & 255 & 2.25      \\
    Official~Language   & 192                         & 3.68                                 & 162 & 2.72      \\
    \bottomrule
  \end{tabular}
\end{table}

%% file: tables/model_inference_specs.tex
\begin{table*}[t]
  \centering
  \scriptsize
  \setlength{\tabcolsep}{3pt}
  \caption[Evaluated model catalogue across 31 checkpoints]{Evaluated model catalogue across all 31 checkpoints with exact Hugging Face repository identifiers.}
  \label{tab:model-inference-specs}
  \begin{tabular*}{\textwidth}{@{\extracolsep{\fill}}lllrrll@{}}
    \toprule
    Family & Stage & Hugging Face Checkpoint & Params & Vocab & Pretraining & Eval Dtype \\
    \midrule
    \textbf{Pythia}      & Pretrained   & \texttt{EleutherAI/pythia-\{160m\dots 12b\}}                    & 0.16B--12B  & 50,304  & 300B (The Pile) & \texttt{float32}/\texttt{16} \\
    \textbf{GPT-2}       & Pretrained   & \texttt{openai-community/gpt2\{,-medium,-large,-xl\}}           & 0.12B--1.5B & 50,257  & WebText         & \texttt{float32} \\
    \textbf{Qwen3-Base}  & Pretrained   & \texttt{Qwen/Qwen3-\{0.6B\dots 14B\}-Base}                      & 0.6B--14B   & 151,936 & 36T             & \texttt{bfloat16} \\
    \textbf{Qwen3}       & Post-trained & \texttt{Qwen/Qwen3-\{0.6B\dots 32B\}}                           & 0.6B--32B   & 151,936 & 36T             & \texttt{bfloat16} \\
    \textbf{Ministral-3} & Base         & \texttt{mistralai/Ministral-3-\{3B,8B,14B\}-Base-2512}          & 3B, 8B, 14B & 131,072 & 1T--3T          & \texttt{bfloat16} \\
    \textbf{Ministral-3} & Instruct     & \texttt{mistralai/Ministral-3-\{3B,8B,14B\}-Instruct-2512}      & 3B, 8B, 14B & 131,072 & 1T--3T          & \texttt{bfloat16} \\
    \textbf{Ministral-3} & Reasoning    & \texttt{mistralai/Ministral-3-\{3B,8B,14B\}-Reasoning-2512}     & 3B, 8B, 14B & 131,072 & 1T--3T          & \texttt{bfloat16} \\
    \bottomrule
  \end{tabular*}
\end{table*}